\documentclass[runningheads]{llncs}
\RequirePackage{silence}  
\usepackage{graphicx}
\usepackage{comment}
\usepackage{amsmath,amssymb}
\usepackage{color}
\usepackage{url}
\usepackage{hyperref}
\usepackage{booktabs}
\usepackage{tabularx}
\usepackage{siunitx}
\usepackage{multirow}
\usepackage{adjustbox}
\usepackage[table]{xcolor} %
\usepackage{array}      %
\usepackage{siunitx}

\usepackage{tikz}
\usepackage{fontawesome}
\usetikzlibrary{positioning}
\usetikzlibrary{shapes.geometric, shapes, arrows, arrows.meta, bending}
\usetikzlibrary{spy}
\usetikzlibrary{backgrounds}
\usepackage{pgfplots}
\usepackage{calc}
\pgfplotsset{compat=1.17}
\usepgfplotslibrary{groupplots}

\usepackage{orcidlink}

\usepackage[capitalize]{cleveref}
\crefname{section}{Sec.}{Secs.}
\Crefname{section}{Section}{Sections}
\crefname{table}{Tab.}{Tabs.}
\Crefname{table}{Table}{Tables}

\definecolor{tabfirst}{rgb}{1, 0.7, 0.7} %
\definecolor{tabsecond}{rgb}{1, 0.85, 0.7} %

\definecolor{tud0d}{RGB}{83,83,83}
\definecolor{tud0c}{RGB}{137,137,137}
\definecolor{tud0b}{RGB}{181,181,181}
\definecolor{tud0a}{RGB}{220,220,220}
\definecolor{tud1a}{RGB}{93,133,195}
\definecolor{tud2a}{RGB}{0,156,218}
\definecolor{tud3a}{RGB}{80,182,149}
\definecolor{tud4a}{RGB}{175,204,80}
\definecolor{tud5a}{RGB}{221,223,72}
\definecolor{tud6a}{RGB}{255,224,92}
\definecolor{tud7a}{RGB}{248,186,60}
\definecolor{tud8a}{RGB}{238,122,52}
\definecolor{tud9a}{RGB}{233,80,62}
\definecolor{tud10a}{RGB}{201,48,142}
\definecolor{tud11a}{RGB}{128,69,151}
\definecolor{tud1b}{RGB}{0,90,169}
\definecolor{tud2b}{RGB}{0,131,204}
\definecolor{tud3b}{RGB}{0,157,129}
\definecolor{tud4b}{RGB}{153,192,0}
\definecolor{tud5b}{RGB}{201,212,0}
\definecolor{tud6b}{RGB}{253,202,0}
\definecolor{tud7b}{RGB}{245,163,0}
\definecolor{tud8b}{RGB}{236,101,0}
\definecolor{tud9b}{RGB}{230,0,26}
\definecolor{tud10b}{RGB}{166,0,132}
\definecolor{tud11b}{RGB}{114,16,133}
\definecolor{tud1c}{RGB}{0,78,138}
\definecolor{tud2c}{RGB}{0,104,157}
\definecolor{tud3c}{RGB}{0,136,119}
\definecolor{tud4c}{RGB}{127,171,22}
\definecolor{tud5c}{RGB}{177,189,0}
\definecolor{tud6c}{RGB}{215,172,0}
\definecolor{tud7c}{RGB}{210,135,0}
\definecolor{tud8c}{RGB}{204,76,3}
\definecolor{tud9c}{RGB}{185,15,34}
\definecolor{tud10c}{RGB}{149,17,105}
\definecolor{tud11c}{RGB}{97,28,115}
\definecolor{tud1d}{RGB}{36,53,114}
\definecolor{tud2d}{RGB}{0,78,115}
\definecolor{tud3d}{RGB}{0,113,94}
\definecolor{tud4d}{RGB}{106,139,55}
\definecolor{tud5d}{RGB}{153,166,4}
\definecolor{tud6d}{RGB}{174,142,0}
\definecolor{tud7d}{RGB}{190,111,0}
\definecolor{tud8d}{RGB}{169,73,19}
\definecolor{tud9d}{RGB}{156,28,38}
\definecolor{tud10d}{RGB}{115,32,84}
\definecolor{tud11d}{RGB}{76,34,106}

\definecolor{myyellow}{RGB}{251,231,163}
\definecolor{myred}{RGB}{246,199,198}

\newcommand \colorindicator[1]{%
	{\textcolor{#1}{$\blacksquare\!\!\!\!\!\blacksquare$}}%
}
\newcommand{\methodname}{CEFITO}

\newif\ifreview
\reviewfalse

\ifreview
	\usepackage{lineno}

	\linenumbers
\fi

\makeatletter
\let\titleold\title
\renewcommand{\title}[1]{\titleold{#1}\newcommand{\thetitle}{#1}}
\def\maketitlesupplementary
   {
   \newpage
        {\centering
        \Large
        \textbf{\thetitle}\\
        \vspace{0.5em}Supplementary Material \\
        \vspace{1.0em}
        }
    }

\usepackage{placeins}

\begin{document}

\def\SubNumber{---}

\def\GCPRTrack{Main Track}

\title{Contrastive Energy Fields for Inference-Time Procedure Planning in Instructional Videos}

\ifreview
	\titlerunning{GCPR 2026 Submission \SubNumber{}. CONFIDENTIAL REVIEW COPY.}
	\authorrunning{GCPR 2026 Submission \SubNumber{}. CONFIDENTIAL REVIEW COPY.}
	\author{GCPR 2026 - \GCPRTrack{}}
	\institute{Paper ID \SubNumber}
\else
	\titlerunning{Contrastive Energy Fields for Inference-Time Procedure Planning}

	\author{Mohamed Afham\inst{1,3\,}\orcidlink{0000-0002-5767-9566} \and
	Christoph Reich\inst{1,2,3,4\,}\orcidlink{0000-0002-8616-1627} \and
	Oliver Hahn\inst{1\,}\orcidlink{0009-0008-6164-1035} \and\\
    Daniel Cremers\inst{2,3,4\,}\orcidlink{0000-0002-3079-7984} \and
    Stefan Roth\inst{1,3,5\,}\orcidlink{0000-0001-9002-9832}}
	
	\authorrunning{M. Afham \emph{et al.}}

    \institute{
        \textsuperscript{1}\,TU Darmstadt\;\;\,
        \textsuperscript{2}\,TU Munich\;\;\,
        \textsuperscript{3}\,ELIZA\;\;\,
        \textsuperscript{4}\,MCML\;\;\,
        \textsuperscript{5}\,hessian.AI\\
        \email{afham.aflal@visinf.tu-darmstadt.de}\\
        \url{https://visinf.github.io/cefito}
    }
    
\fi

\maketitle              %

\begin{abstract}
Procedure planning seeks to estimate a sequence of actions to transition from an observed initial state to a given goal state. Current procedure planning approaches directly predict action sequences from latent representations using feed-forward neural networks or diffusion-based inference. These paradigms treat every action as plausible, lacking the ability to enforce task-specific logical constraints that render certain actions irrelevant or not plausible. We propose \methodname{}, a procedure planning approach that learns a predictor to express an action-conditioned representation space. Based on this representation space, we formulate procedure planning as a task-constrained optimization problem. Unlike prior methods, \methodname{} explicitly reasons over the action space by omitting irrelevant actions during inference-time planning. This reformulation enables effective procedure planning and achieves state-of-the-art accuracy on two established procedure planning benchmarks.%
\end{abstract}
\section{Introduction} \label{sec:introduction}

\begin{figure}[t]
    \centering
    \def\svgwidth{1.0\textwidth}
    \input{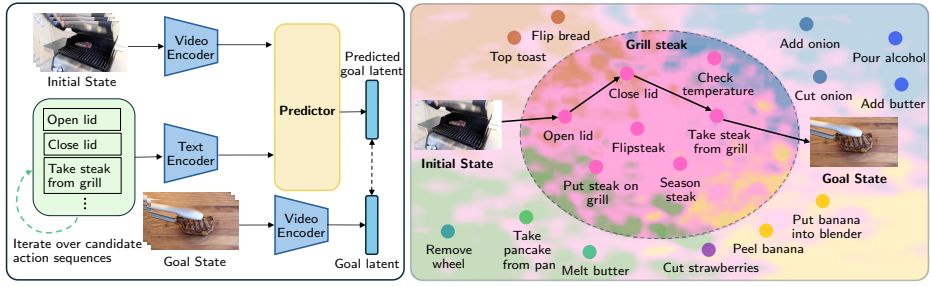}
    \vspace{-1.5em}
    \caption{\textbf{Overview of \methodname{}.} \emph{Left:} Given an initial observation and a sequence of actions, our method maps action sequences to a predicted goal latent. The distance to the encoded goal latent defines the candidate's energy, which we minimize at inference. \emph{Right:} We learn a task-conditioned energy field via contrastive learning. At inference, \methodname{} predicts the task (\emph{e.g.}, ``Grill steak''), restricting the search to an action subset (pink region). The action sequence with minimal energy is the predicted plan.}   
    \label{fig:teaser}
\end{figure}

Planning is a central component of human intelligence \cite{Miller:1960:PLN}. Humans can reason about the next actions to take and their consequences in order to reach a desired goal. The ability to plan is crucial in real-world tasks, including robotic navigation \cite{nwm,govig}, autonomous driving \cite{hu2023planning}, virtual reality \cite{queisner2024surgical}, and healthcare~\cite{xu2026healt}. Procedure planning in instructional videos~\cite{ddn} seeks to plan a sequence of actions given a visual initial and a desired goal state. Unlike classical planning formulations with explicit initial and goal states, procedure planning in instructional videos directly operates on raw visual observation (\emph{i.e.}, imagery), providing a vision-driven planning setting grounded in real-world environments~\cite{coin,crosstask}.

Current approaches from procedure planning employ various deep learning architectures, including transformers~\cite{plate}, diffusion models~\cite{pdpp,mtid}, and task-specific procedural knowledge graphs \cite{vitrebiplannet}. Still, these approaches use feed-forward or diffusion-based inference to map latent representations to a distribution over action sequences. In contrast, human cognitive planning is a dynamic and iterative process \cite{planningasinference,cogplan}. Humans do not predict a single sequence. Instead, humans infer a set of possible actions and optimize over these to reach a desired goal. Current procedure planning approaches do not mimic this process and cannot, for example, disregard irrelevant actions during inference.

Instead of feed-forward inference, work in classical robot control and planning performs online optimization using predictive models \cite{Camacho:2007:MPC}. More recently, action-conditioned world models have demonstrated effective planning using inference-time optimization~\cite{worldmodel,lecun2022path,le-wm}. During training, a predictive model is learned to express the transition from a current state to the next. At inference, this predictive model is used for planning by optimizing over a sequence of actions. Such approaches capture the dynamics of the underlying system and the relationship between actions and their outcomes. In procedure planning, modeling action-conditioned visual transitions is crucial, yet challenging due to the scarcity of training data, including intermediate observations. Existing approaches use Large Language Models \cite{schema,planllm}, interpolation \cite{mtid}, or probabilistic graphs \cite{kepp,vitrebiplannet} as proxies for unknown intermediate states. However, these approaches rely on rigid structures or require huge training datasets and large-scale pre-training.

Motivated by the effectiveness of inference-time optimization using predictive models, we introduce \methodname{}---\textbf{C}ontrastive \textbf{E}nergy \textbf{F}ields for \textbf{I}nference-\textbf{T}ime \textbf{O}ptimization in Procedure Planning in Instructional Videos. In contrast to existing procedure planning approaches, \methodname{} (\emph{cf.} \cref{fig:teaser}) decomposes procedure planning into two sub-problems. \emph{First}, \methodname{} learns an action-conditioned representation space (\emph{i.e.}, an energy field) in the form of a predictor model, using contrastive learning. This predictor models the transition from an initial state to a goal state, given a sequence of actions. \emph{Second}, at inference-time, we formulate the planning of an action sequence as a task-constrained optimization problem using our predictor model. Unlike prior approaches, \methodname{} can explicitly reason over the action space, allowing for the omission of irrelevant actions during planning. This allows \methodname{} to perform effective and accurate procedure planning in instructional videos.

Specifically, we make the following contributions: 
 \emph{(i)} We formulate procedure planning as energy minimization over an action-conditioned representation, learned with a contrastive objective that does not require intermediate state supervision.
\emph{(ii)} We introduce a task-constrained search procedure that exploits this energy to plan over a restricted, task-relevant action subset at inference time.
\emph{(iii)} We evaluate \methodname{} on established procedure planning benchmarks of instructional videos. \methodname{} achieves state-of-the-art accuracy on two datasets and different planning horizons, while not relying on large language models.

\section{Related Work} \label{sec:related_work}

\subsubsection{Procedure Planning in Instructional Videos.} Procedure planning in instructional videos~\cite{ddn} aims to plan a sequence of actions transitioning from an initial state to a desired goal state. Both the initial and the goal state are given as visual observations (\emph{i.e.}, video frames). This task is closely related to general task planning~\cite{ddn,Ghallab:2004:PLAN}. Initial approaches for procedure planning employed sequence modeling via sequential latent spaces~\cite{ddn} or adversarial policy learning~\cite{extgail}. While DDN~\cite{ddn} and Ext-GAIL~\cite{extgail} learn using sequential action supervision, subsequent works employed additional supervision such as language~\cite{e3p,p3iv}. In particular, P\textsuperscript{3}IV~\cite{p3iv} explores supervision with natural language representations of the action labels instead of one-hot labels, while EGPPP~\cite{e3p}, in addition, leverages task labels to condition the planning model. PlaTe~\cite{plate} introduced a transformer-based model, reducing compounding prediction errors common with single-step models. PDPP \cite{pdpp} and MTID \cite{mtid} employ diffusion to probabilistically model the prediction action sequences. KEPP \cite{kepp} leverages a probabilistic procedural knowledge graph to guide state transition. Building on this, ViterbiPlanNet~\cite{vitrebiplannet} introduced a differentiable viterbi layer upon the probabilistic knowledge graph to enable end-to-end training. A different line of work approached procedure planning using large language models (LLMs) \cite{schema,planllm}. SCHEMA \cite{schema} uses an LLM to explicitly model state transitions, while PlanLLM \cite{planllm} fine-tunes an LLM and trains a step decoder. Different from these approaches that utilize feed-forward neural networks or diffusion models, we learn an action-conditioned representation space using contrastive learning. This allows us to express an energy function and perform task-constrained optimization-based inference.

\subsubsection{Inference-Time Optimization for Planning.} Inference-time optimization is a well-established paradigm in robotics control and planning, where action sequences are optimized online using forward predictive models \cite{Camacho:2007:MPC}. Early visual foresight methods established this direction by combining pixel-level rollouts or learned dynamics with model predictive control to plan trajectories at inference time, without updating policy parameters during inference~\cite{visual-foresight-ebert,foresight,nagabandi}. The introduction of world models~\cite{vjepav2,worldmodel,lecun2022path} has further advanced this idea by moving from high-dimensional pixel space to latent representations, enabling online trajectory optimization at inference time~\cite{hafner,janner}. More recently, large-scale visual foundation models~\cite{vjepav2,dino,dinov2} have been used to build world models that support planning entirely during inference~\cite{nwm,le-wm,dino-wm}. Motivated by classical methods, we reformulate procedure planning as an inference-time optimization problem using an action-conditioned energy field.

\subsubsection{Contrastive Learning.} Contrastive learning aims to learn expressive representations in a self-supervised \cite{swav,simclr}, weakly \cite{clip-orig}, or supervised fashion \cite{supcon}. At its core, contrastive learning enforces discrimination between pairs of positive and negative data points and has been applied for downstream tasks, including action recognition~\cite{KhorasganiCS22,avid,WangBTT22}, action anticipation~\cite{QiWSSHT23}, and temporal action localization~\cite{GaoCX22,JuZLZZC0W23}. A central aspect for the effectiveness of contrastive learning is the choice and number of negative samples~\cite{hard-neg}. Specifically, hard negatives, which are semantically similar to positive samples, provide an informative training signal. In this work, we utilize supervised contrastive learning \cite{supcon} in the form of a contrastive triplet-loss \cite{facenet} to learn an action-conditioned representation space suitable for planning at inference time.

\section{Method: \methodname{}} \label{sec:method}

In this section, we present \methodname{} for procedure planning in instructional videos. We will first revisit the task definition of procedure planning and reformulate the problem as energy minimization over an action-conditioned energy field (\emph{cf.} \cref{sec:pptoito}). Next, we introduce a contrastive learning approach for obtaining this action-conditioned energy field in the form of a predictor model (\emph{cf.} \cref{sec:contrastive}). Finally, we propose our task-constrained inference-time optimization for planning action sequences (\emph{cf.} \cref{sec:inference}).

\subsection{Procedure Planning as Inference-Time Optimization}\label{sec:pptoito}

\subsubsection{Problem Formulation.} Procedure planning seeks to estimate a sequence of actions $a_{1:\rm T}$ from a predefined corpus of actions, transforming a given initial state $v_s$ into a given goal state $v_g$. $\rm T$ represents the planning horizon, \emph{i.e.}, the number of action steps required to achieve the goal state $v_g$. Both the initial state and the goal state are provided as visual observations (\emph{i.e.}, video frames). Following the established paradigm \cite{ddn,kepp,schema,plate,pdpp,planllm,mtid}, we encode the visual initial $v_s$ and goal state $v_g$ using a visual feature extractor (\emph{i.e.}, S3D~\cite{endtoend} pre-trained on the HowTo100M dataset \cite{howto100m}), resulting in the latent initial state $x_s\in\mathbb{R}^{E}$ and the latent goal state $x_g\in\mathbb{R}^{E}$. $E$ denotes the latent dimension of the visual features.

\subsubsection{Reformulating Procedure Planning.} Prior approaches map $(x_s, x_g)$ directly to an action sequence through a feed-forward or diffusion-based network \cite{plate,pdpp,mtid}. We reformulate procedure planning as an inference-time search over a learned action-conditioned energy field, decomposing it into two sub-tasks. \emph{First}, we learn a predictor model that maps a candidate action sequence and the initial state $x_s$ to the resulting goal embedding, acting as an energy. \emph{Second}, at inference, we optimize over candidate action sequences and select the one whose predicted goal embedding is closest to the observed latent state $x_g$.

\begin{figure}[t]
    \centering
    \def\svgwidth{1.0\textwidth}
    \input{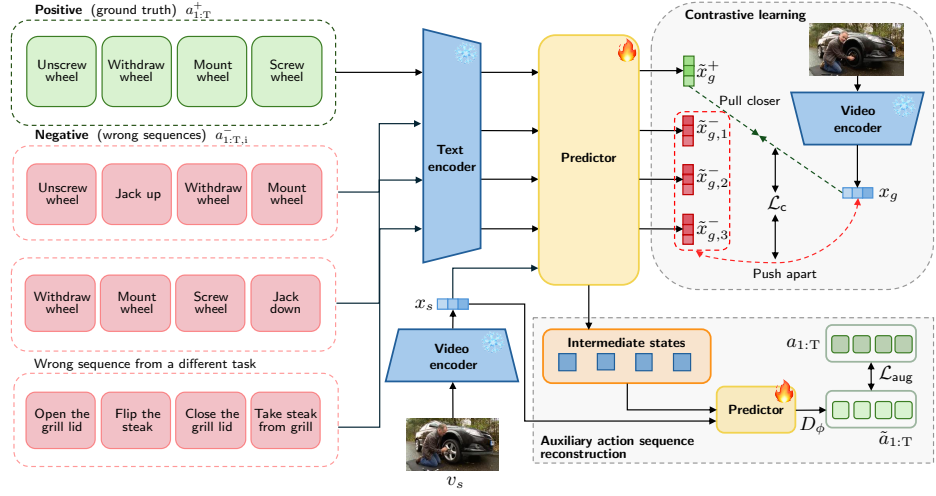}
    \vspace{-1.5em}
    \caption{\textbf{\methodname{} training framework.} During training, a positive and $N$ negative (wrong) action sequences are provided. Each sequence is tokenized using a text encoder and fed into the predictor $P_{\theta}$, alongside the initial latent state $x_s$. For each sequence we predict a goal state $\tilde{x}_g$, while predictions of negative sequences $\tilde{x}_{g,i}^{-}$ are pushed away from the target $\tilde{x}_g$ and for positive sequences $\tilde{x}_g^{+}$ pulled together. An auxiliary loss $\mathcal{L}_{\text{aux}}$ supervises auxiliary latents of $P_{\theta}$ via a decoding head $D_{\phi}$.}
    \label{fig:architecture_training}
\end{figure}

\subsection{Action-Conditioned Energy Field}\label{sec:contrastive}

To perform planning at inference time, we aim to learn an action-conditioned energy field capturing the transition from $x_s$ to $x_g$, given an action sequence. Perpetually, we train a predictor model $P_{\theta}$, mapping an action sequence $a_{1:\rm T}$ (textual form) and an initial state $x_s$ to a predicted goal state $\tilde{x}_g\in\mathbb{R}^{E}$ by
\begin{equation}\label{eq:predictor}
    \tilde{x}_g = P_{\theta}(x_s, a_{1:\rm T}).
\end{equation}
The objective of $P_{\theta}$ is to predict $\tilde{x}_g$ close to the observed goal $x_g$ when conditioned on the correct action sequence $a^{+}_{1:T}$, and far from $x_g$ for any incorrect sequence $a^{-}_{1:T}$.
During inference, we can search for an action sequence that minimizes distance in latent space. Formally, $d\!\left(P_{\theta}(x_s, a_{1:\rm T}), x_g\right)$ can be seen as an energy function, where $d(\cdot,\cdot)$ denotes a distance metric.

We implement our predictor model $P_{\theta}$ as a transformer \cite{transformer}. We use the initial latent state $x_s$ as an input token. The action sequence is encoded in language form using CLIP \cite{clip-orig}. The resulting embeddings are used as additional input tokens. We visualize the predictor architecture in \cref{fig:architecture_training}.

\subsubsection{Contrastive Training.} We train our predictor $P_{\theta}$ using contrastive learning. In particular, we use a margin-based triplet loss~\cite{facenet}. 
Given the \emph{correct} action sequence $a^{+}_{1:\rm T}$, we obtain $\tilde{x}^{+}_g\in\mathbb{R}^{E}$ using $P_{\theta}$ and $x_s$ (\emph{cf.} \cref{eq:predictor}). Analogously, given $N$ \emph{wrong} sequences $\{a^{-}_{1:\rm T, 1}, \ldots, a^{-}_{1:\rm T, N}\}$, we obtain $\tilde{x}^{-}_{g,i}\in\mathbb{R}^{C\times N}$. Given the ground truth latent goal state, we compute our contrastive loss $\mathcal{L}_{\text{c}}$ using
\begin{equation}\label{eq:contrastiveloss}
    \mathcal{L}_{\text{c}}=\frac{1}{N}\sum_{i=1}^N \max\!\left(\left\Vert x_{g} - \tilde{x}_g^{+}\right\Vert_{2} - \left\Vert x_{g} - \tilde{x}_{g,i}^{-}\right\Vert_{2} + \tau_i, 0\right).
\end{equation}
Here, $\tau_{i}\in\mathbb{R}^{+}$ is the triplet margin per negative sample $i$, enforcing a minimum separation and preventing the degenerate solution that collapses all embeddings. We train our predictor's weights $\theta$ to minimize \cref{eq:contrastiveloss}.

\paragraph{Negative Action Sequence Sampling.} Negative samples are critical for obtaining a discriminative space \cite{neg-sample-mixing,hard-neg}. To learn an energy field that captures both high-level (\emph{e.g.}, discriminate between tasks) and low-level structure (\emph{e.g.}, detect wrong action sequence order), we employ mixed-negative sampling. By generating both hard and easy negative samples, we impose high and low-level structure.

Given a ground truth action sequence with initial, goal state, and task label $c \in \mathcal{C}$, we generate $N$ negative samples, \emph{i.e.}, wrong sequences. Specifically, $r\,N$ of the negative samples are hard negatives and $(1-r)\,N$ are easy negatives. $r\in[0, 1]$ is the hard/easy negative sample ratio, controlling the balance between both. Hard negative samples share the same high-level task $c$ but differ in ordering or composition. Easy negatives are sequences sampled from tasks other than $c$.

\paragraph{Adaptive Margins.} The triplet-loss margin $\tau_i$ (\emph{cf.} \cref{eq:contrastiveloss}) enforces a minimum distance between the positive $\left\Vert x_{g} - x_g^{+}\right\Vert_{2}$ and negative pair $\left\Vert x_{g} - x_{g,i}^{-}\right\Vert_{2}$. 
Using a fixed margin for every negative is suboptimal because negatives differ in similarity to the ground truth:
An action sequence only differing from the ground truth in the ordering of the actions (\emph{i.e.}, hard negative sample) should ideally entail a smaller distance than an action sequence of unrelated actions (\emph{i.e.}, easy negative sample). To enforce this structure, we use adaptive margins, assigning a margin to each negative sample based on the action overlap with the positive sample. In particular, given the positive action sequence $a^{+}_{1:\rm T}$ (\emph{i.e.}, ground truth) and a negative action sequence $a^{-}_{1:\rm T, i}$, we extract the sets of unique actions of the positive $\mathbb{A}^{+}$ and of the negative sequence $\mathbb{A}^{-}_{i}$. Next, we compute $\tau_i$ using
\begin{equation}
    \tau_i = \tau_{\text{min}} + (\tau_{\text{max}} - \tau_{\text{min}})\frac{|\mathbb{A}^{-}_{i}\setminus\mathbb{A}^{+}|}{|\mathbb{A}^{-}_{i}|}.
\end{equation}
Here, the hyperparameters $\tau_{\text{min}}\in\mathbb{R}^{+}$ denotes the minimum and $\tau_{\text{max}}\in\mathbb{R}^{+}$ the maximum margin, bounding the per negative margin $\tau_i\in[\tau_{\text{min}}, \tau_{\text{max}}]$.

\subsubsection{Auxiliary Action Sequence Reconstruction} While \cref{eq:contrastiveloss} supervises the final output representation, we further enforce expressive intermediate representations using an action sequence reconstruction approach. As visual representations of the intermediate states are not available, we employ a mask token modeling \cite{devlin2019bert,he2022masked} approach. 
We feed the initial latent state $x_{s}$ and the action sequence $a_{1:\rm T}$ (tokenized using the text encoder) to our predictor $P_{\theta}$. Additionally, while omitted in \cref{eq:predictor} for clarity, we also feed a learnable masked token for each action into the predictor. The predictor outputs the goal state representation $\tilde{x}_{g}$, we further extract the intermediate representation $\{\tilde{x}_{l}\}^{\rm T}_{t=1}$ corresponding to each of the masked tokens. After obtaining $\{\tilde{x}_{l}\}^{\rm T - 1}_{t=1}$, now reverse the masking and all actions $a_{1:\rm T}$ and feed $\{\tilde{x}_{l}\}^{\rm T - 1}_{t=1}$ into our predictor and a decoding head $D_{\phi}$. This yields a reconstruction of actions $\tilde{a}_{1:\rm T}$. Each reconstructed action is supervised using the ground-truth actions $a_{1:\rm T}$ using $\mathcal{L}_{\text{aux}} = \frac{1}{\rm T} \sum_{k=1}^{\rm T}\mathcal{L}_{\text{CE}}\!\left(\tilde{a}_{k}, a_{k}\right)$. Here, $\mathcal{L}_{\text{CE}}$ denotes the cross-entropy loss, and actions are mapped to the vocabulary of actions. We omit this mapping for the sake of compactness.

Our full predictor loss composes both the contrastive loss $\mathcal{L}_{\text{c}}$ and our auxilary loss $\mathcal{L}_{\text{aux}}$, weighted by $\lambda$, \emph{i.e.}, $\mathcal{L} = \mathcal{L}_{\text{c}} + \lambda\, \mathcal{L}_{\text{aux}}$. We visualize our training in \cref{fig:architecture_training}.

\begin{figure}[t]
    \centering
    \def\svgwidth{1.0\textwidth}
    \input{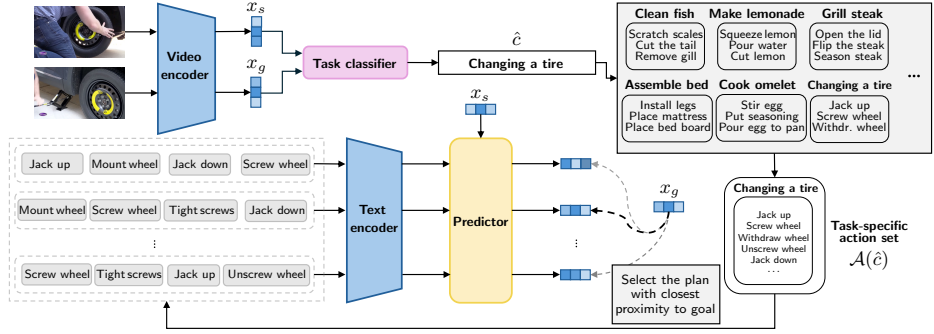}
    \vspace{-1.5em}
    \caption{\textbf{\methodname{} inference.} A task classifier predicts $\hat{c}$ from $(x_s, x_g)$ and restricts the search to $\mathcal{A}(\hat{c})$. The predictor $P_{\theta}$ scores every candidate sequence in $\mathcal{A}(\hat{c})$. The sequence approximating $x_g$ best, forms the action sequence prediction $\tilde{a}_{1:\rm T}$.}
    \label{fig:architecture_inference}
\end{figure}

\subsection{Task-Constrained Optimization for Planning}\label{sec:inference}

Equipped with our predictor $P_{\theta}$, we now seek to infer an action sequence, given the initial $x_s$ and goal state $x_g$, in latent space (\emph{cf.} \cref{fig:architecture_inference}). We formulate inference as a discrete search over the action space $\mathcal{A}$. Optimizing over the full space is expensive. To this end, we learn a lightweight classifier that estimates the high-level task $c\in\mathcal{C}$. Implemented as a multilayer perceptron, the classifier takes in $(x_s, x_g)$, and predicts the task $\hat{c}\in\mathcal{C}$. We train the classifier using cross-entropy.

Using the estimated high-level task $\hat{c}$, we reduce the search space from the full action space to a task-specific subspace $\mathcal{A}(\hat{c})$. This subspace only captures actions relevant to the specific task and is derived from the training data. Using $\mathcal{A}(\hat{c})$, the initial state $x_s$, and goal state $x_g$, we infer an action sequence $\tilde{a}_{1:\rm T}$ by
\begin{equation}\label{eq:inference}
    \tilde{a}_{1:\rm T}=\underset{\tilde{a}_{1:\rm T} \in \mathcal{A}(\hat{c})}{\arg\min}\!\left\Vert P_{\theta}(x_s, \tilde{a}_{1:\rm T}) - x_g\right\Vert_{2}.
\end{equation} 
This inference approach enables procedure planning by considering only relevant actions, ignoring irrelevant ones. Our inference approach is shown in \cref{fig:architecture_inference}.

\begin{table*}[t]
\centering
\renewcommand{\arraystretch}{0.975}
\caption{\textbf{Results on CrossTask~\cite{crosstask}.} We compare \methodname{} with the state of the art on CrossTask val., using different planning horizons, and report SR, mACC, and mIoU (all in \%,\,$\uparrow$). Best results highlighted in red \colorindicator{tabfirst}; second-best results in orange \colorindicator{tabsecond}. For completeness, we report baselines not adhering to the protocol by \cite{vitrebiplannet} in gray \colorindicator{gray!70}.}
\setlength{\tabcolsep}{1.535pt}
\scriptsize
\sisetup{table-number-alignment=center}
\begin{tabularx}{\linewidth}{@{}l S[table-format=2.2]lS[table-format=2.2]lS[table-format=2.2]l l S[table-format=2.2]lS[table-format=2.2]lS[table-format=2.2]l@{}}
\toprule
\raisebox{-3.0pt}[0pt][0pt]{\multirow{2}{*}{\textbf{Method}}} 
& \multicolumn{6}{c}{\textbf{T=3}} 
& \hphantom{4} & \multicolumn{6}{c}{\textbf{T=4}} \\
\cmidrule(lr){2-7} \cmidrule(lr){9-14}
& \multicolumn{2}{c}{{SR\,$\uparrow$}} & \multicolumn{2}{c}{mAcc\,$\uparrow$} & \multicolumn{2}{c}{mIoU\,$\uparrow$} 
& & \multicolumn{2}{c}{SR\,$\uparrow$} & \multicolumn{2}{c}{mAcc\,$\uparrow$} & \multicolumn{2}{c}{mIoU\,$\uparrow$} \\
\midrule

WLTDO~\cite{wltdo}
& \color{gray!70} 1.87 & & \color{gray!70} 21.64 & & \color{gray!70} 31.70 & &
& \color{gray!70} 0.77 & & \color{gray!70} 17.92 & & \color{gray!70} 26.43 & \\

UAAA~\cite{uaa}
& \color{gray!70} 2.15 & & \color{gray!70} 20.21 & & \color{gray!70} 30.87 & &
& \color{gray!70} 0.98 & & \color{gray!70} 19.86 & & \color{gray!70} 27.09 & \\

UPN~\cite{upn}
& \color{gray!70} 2.89 & & \color{gray!70} 24.39 & & \color{gray!70} 31.56 & &
& \color{gray!70} 1.19 & & \color{gray!70} 21.59 & & \color{gray!70} 27.85 \\

DDN~\cite{ddn}
& \color{gray!70} 12.18 & & \color{gray!70} 31.29 & & \color{gray!70} 47.48 & &
& \color{gray!70} 5.97 & & \color{gray!70} 27.10 & & \color{gray!70} 48.46 & \\

PlaTe~\cite{plate}
& \color{gray!70} 16.00 & & \color{gray!70} 36.17 & & \color{gray!70} 65.91 & &
& \color{gray!70} 14.00 & & \color{gray!70} 35.29 & & \color{gray!70} 55.36 & \\

Ext-GAIL~\cite{extgail}
& \color{gray!70} 21.27 & & \color{gray!70} 49.46 & & \color{gray!70} 61.70 & &
& \color{gray!70} 16.41 & & \color{gray!70} 43.05 & & \color{gray!70} 60.93 & \\

P$^3$IV~\cite{p3iv}
& \color{gray!70} 23.34 & & \color{gray!70} 49.96 & & \color{gray!70} 73.89 & &
& \color{gray!70} 13.40 & & \color{gray!70} 44.16 & & \color{gray!70} 70.01 & \\

EGPP~\cite{e3p}
& \color{gray!70} 26.40 & & \color{gray!70} 53.02 & & \color{gray!70} 74.05 & &
& \color{gray!70} 16.49 & & \color{gray!70} 48.00 & & \color{gray!70} 70.16 & \\

\midrule

Qwen2.5-VL-32B~\cite{qwen-2.5-vl}
& 11.48 & & 36.35 & & 69.52 & &
& 5.56 & & 31.22 & & 66.31 & \\

Qwen2.5-32B~\cite{qwen-2.5}
& 25.14 & & 56.10 & & 80.92 & &  
& 9.22 & & 46.32 & & 76.15 & \\

Gemini 2.5 Pro~\cite{gemini-2.5}
& 29.18 & & 57.90 & & 81.48 & &    
& 14.00 & & 51.33 & & 78.58 & \\

Qwen3-30B~\cite{qwen3}
& 23.37 & & 55.96 & & 81.16 & &      
& 10.59 & & 49.06 & & 78.03 & \\

Qwen3-30B + PKG~\cite{vitrebiplannet}  
& 23.31 & & 56.15 & & 81.06 & &       
& 10.96 & & 48.77 & & 77.48 & \\

PKG beam search~\cite{vitrebiplannet}
& 22.38 & {\tiny$\pm$\num{0.26}} 
& 55.74 & {\tiny$\pm$\num{0.25}} 
& 80.92 & {\tiny$\pm$\num{0.26}} &
& 9.30 & {\tiny$\pm$\num{0.22}} 
& 47.65 & {\tiny$\pm$\num{0.54}} 
& 78.25 & {\tiny$\pm$\num{0.42}} \\

PDPP~\cite{pdpp}
& 36.73 & {\tiny$\pm$\num{0.59}} 
& 61.96 & {\tiny$\pm$\num{0.59}} 
& 83.20 & {\tiny$\pm$\num{0.33}} &
& 21.47 & {\tiny$\pm$\num{2.09}} 
& 55.66 & {\tiny$\pm$\num{1.64}} 
& 80.68 & {\tiny$\pm$\num{0.83}} \\

KEPP~\cite{kepp}
& 34.93 & {\tiny$\pm$\num{2.60}} 
& 60.34 & {\tiny$\pm$\num{1.61}} 
& 82.67 & {\tiny$\pm$\num{0.69}} &
& 22.34 & {\tiny$\pm$\num{0.43}} 
& 55.24 & {\tiny$\pm$\num{0.30}} 
& 80.58 & {\tiny$\pm$\num{0.25}} \\

PlanLLM~\cite{planllm}
& 36.84 & {\tiny$\pm$\num{1.21}} 
& 61.56 & {\tiny$\pm$\num{1.03}} 
& 83.23 & {\tiny$\pm$\num{0.53}} &
& 22.91 & {\tiny$\pm$\num{1.39}} 
& 55.29 & {\tiny$\pm$\num{1.54}} 
& 81.03 & {\tiny$\pm$\num{0.47}} \\

SCHEMA~\cite{schema}
& 37.24 & {\tiny$\pm$\num{0.60}} 
& 62.69 & {\tiny$\pm$\num{0.28}} 
& \cellcolor{tabsecond} 83.94 & \cellcolor{tabsecond}{\tiny$\pm$\num{0.23}} &
& 24.18 & {\tiny$\pm$\num{0.47}} 
& \cellcolor{tabsecond} 57.02 & \cellcolor{tabsecond}{\tiny$\pm$\num{0.64}} 
& \cellcolor{tabsecond} 81.46 & \cellcolor{tabsecond}{\tiny$\pm$\num{0.19}} \\

ViterbiPlanNet~\cite{vitrebiplannet}
& \cellcolor{tabsecond} 38.45 & \cellcolor{tabsecond}{\tiny$\pm$\num{0.32}} 
& \cellcolor{tabsecond} 63.07 & \cellcolor{tabsecond}{\tiny$\pm$\num{0.17}} 
& 83.89 & {\tiny$\pm$\num{0.16}} &
& \cellcolor{tabsecond} 24.64 & \cellcolor{tabsecond}{\tiny$\pm$\num{0.30}} 
& 57.00 & {\tiny$\pm$\num{0.42}} 
& 81.18 & {\tiny$\pm$\num{0.44}} \\

\midrule
\methodname{} (Ours)
& \cellcolor{tabfirst} 39.62 & \cellcolor{tabfirst}{\tiny$\pm$0.24} 
& \cellcolor{tabfirst} 64.12 & \cellcolor{tabfirst}{\tiny$\pm$0.31} 
& \cellcolor{tabfirst} 84.29 & \cellcolor{tabfirst}{\tiny$\pm$0.21} & \cellcolor{tabfirst}
& \cellcolor{tabfirst} 24.76 & \cellcolor{tabfirst}{\tiny$\pm$0.43} 
& \cellcolor{tabfirst} 57.53 & \cellcolor{tabfirst}{\tiny$\pm$0.37} 
& \cellcolor{tabfirst} 81.58 & \cellcolor{tabfirst}{\tiny$\pm$0.25} \\

\bottomrule
\end{tabularx}%

\label{tab:crosstask}

\end{table*}

\section{Experiments} \label{sec:experiments}

We evaluate \methodname{} on two established procedure planning benchmarks and compare against existing state-of-the-art approaches (\emph{cf.} \cref{sec:mainresults}). Moreover, we analyze the main components and hyperparameters of our approach (\emph{cf.} \cref{sec:analysis}).

\subsubsection{Datasets.} We evaluate on two procedure planning datasets, CrossTask \cite{crosstask} and COIN \cite{coin}. CrossTask contains \num{2750} videos spanning \num{18} high-level tasks with a total of \num{105} distinct actions. Each video entails \num{7.6} actions on average. COIN comprises \num{11827} videos of \num{180} different tasks, covering \num{778} unique actions. We adopt the unified evaluation protocol by Seminara \textit{et al.} \cite{vitrebiplannet}, where we train using five random seeds and report the mean and the \SI{90}{\%} confidence interval.

\subsubsection{Metrics.} We follow existing work \cite{kepp,schema,vitrebiplannet,pdpp,planllm,p3iv,mtid} and utilize three standard metrics to measure procedure planning accuracy. \emph{First}, we utilize the Success Rate (SR), measuring the percentage of predicted action sequences that exactly match the corresponding ground truth sequences. \emph{Second}, we report the Mean Accuracy (mAcc), measuring the average proportion of correctly predicted actions across all step-wise positions. \emph{Third}, we use the Mean Intersection over Union (mIoU), computing the overlap between the predicted and ground truth action sequences. In particular, we follow the element-wise mIoU formulation suggested by Seminara \textit{et al.} \cite{vitrebiplannet}. Among these three metrics, SR is the strictest metric. All metrics are reported in \%.

\begin{table*}[t]
\renewcommand{\arraystretch}{0.975}
\centering
\caption{\textbf{Results on COIN \cite{coin}.} We compare \methodname{} with the state of the art on COIN val., using different planning horizons, and report SR, mACC, and mIoU (all in \%,\,$\uparrow$). Best results are highlighted in red \colorindicator{tabfirst} and second-best results in orange \colorindicator{tabsecond}. For completeness, we report baselines not adhering to the protocol by \cite{vitrebiplannet} in gray \colorindicator{gray!70}.}
\setlength{\tabcolsep}{0.885pt}
\scriptsize
\sisetup{table-number-alignment=center}
\begin{tabularx}{\linewidth}{@{}l S[table-format=2.2]lS[table-format=2.2]lS[table-format=2.2]l l S[table-format=2.2]lS[table-format=2.2]lS[table-format=2.2]l@{}}
\toprule
\raisebox{-3.0pt}[0pt][0pt]{\multirow{2}{*}{\textbf{Method}}} 
& \multicolumn{6}{c}{\textbf{T=3}} 
& \hphantom{1} & \multicolumn{6}{c}{\textbf{T=4}} \\
\cmidrule(lr){2-7} \cmidrule(lr){9-14}
& \multicolumn{2}{c}{{SR\,$\uparrow$}} & \multicolumn{2}{c}{mAcc\,$\uparrow$} & \multicolumn{2}{c}{mIoU\,$\uparrow$} 
& & \multicolumn{2}{c}{SR\,$\uparrow$} & \multicolumn{2}{c}{mAcc\,$\uparrow$} & \multicolumn{2}{c}{mIoU\,$\uparrow$} \\
\midrule

DDN~\cite{ddn}
& \color{gray!70} 13.90 & & \color{gray!70} 20.19 & & \color{gray!70} 64.78 & &
& \color{gray!70} 11.13 & & \color{gray!70} 17.71 & & \color{gray!70} 68.06 & \\

P\textsuperscript{3}IV~\cite{p3iv}
& \color{gray!70} 15.40 & & \color{gray!70} 21.67 & & \color{gray!70} 76.31 & &
& \color{gray!70} 11.32 & & \color{gray!70} 18.85 & & \color{gray!70} 70.53 & \\

EGPP~\cite{e3p}
& \color{gray!70} 19.57 & & \color{gray!70} 31.42 & & \color{gray!70} 84.95 & &
& \color{gray!70} 13.59 & & \color{gray!70} 26.72 & & \color{gray!70} 84.72 & \\

\midrule

Qwen2.5-VL-32B~\cite{qwen-2.5-vl}

& 3.65 & & 17.51 & & 52.10 & &
& 1.87 & & 17.05 & & 55.66 & \\

Qwen2.5-32B~\cite{qwen-2.5}
& 14.97 & & 36.34 & & 78.74 & &  
& 4.98 & & 27.45 & & 71.64 & \\

Gemini 2.5 Pro~\cite{gemini-2.5}
& 17.02 & & 38.87 & & 78.73 & &    
& 8.10 & & 31.90 & & 71.70 & \\

Qwen3-30B~\cite{qwen3}
& 14.52 & & 36.56 & & 78.07 & &   
& 4.64 & & 28.85 & & 70.45 & \\

Qwen3-30B\,+\,PKG~\cite{vitrebiplannet}  
& 14.63 & & 36.53 & & 78.11 & &       
& 4.78 & & 29.00 & & 71.04 & \\

PKG beam search~\cite{vitrebiplannet}
& 13.32 & {\tiny$\pm$\hphantom{1}\num{0.34}} 
& 37.42 & {\tiny$\pm$\hphantom{1}\num{1.19}} 
& 78.93 & {\tiny$\pm$\hphantom{1}\num{2.06}} &
& 5.14 & {\tiny$\pm$\hphantom{1}\num{0.60}} 
& 31.29 & {\tiny$\pm$\hphantom{1}\num{3.64}} 
& 74.26 & {\tiny$\pm$\hphantom{1}\num{5.38}} \\

PDPP~\cite{pdpp}
& 22.37 & {\tiny $\pm$\hphantom{1}0.57}
& 44.60 & {\tiny $\pm$\hphantom{1}0.16}
& 83.00 & {\tiny $\pm$\hphantom{1}0.42} &
& 15.21 & {\tiny $\pm$\hphantom{1}0.34}
& 41.01 & {\tiny $\pm$\hphantom{1}0.32}
& 81.64 & {\tiny $\pm$\hphantom{1}0.48} \\

KEPP~\cite{kepp}
& 13.85 & {\tiny $\pm$\hphantom{1}7.49}
& 28.40 & {\tiny $\pm$12.26}
& 62.54 & {\tiny $\pm$14.35} &
& 15.20 & {\tiny $\pm$\hphantom{1}1.27}
& 33.39 & {\tiny $\pm$\hphantom{1}0.73}
& 67.79 & {\tiny $\pm$\hphantom{1}1.29} \\

PlanLLM~\cite{planllm}
& 33.44 & {\tiny $\pm$\hphantom{1}0.15}
& \cellcolor{tabsecond}51.05 & {\cellcolor{tabsecond}\tiny $\pm$\hphantom{1}0.46}
& \cellcolor{tabsecond}84.66 & {\cellcolor{tabsecond}\tiny $\pm$\hphantom{1}0.41} &
& 23.19 & {\tiny $\pm$\hphantom{1}0.32}
& \cellcolor{tabsecond} 45.70 & {\cellcolor{tabsecond}\tiny $\pm$\hphantom{1}0.33}
& \cellcolor{tabfirst}83.44 & {\cellcolor{tabfirst}\tiny $\pm$\hphantom{1}0.39} \\

SCHEMA~\cite{schema}
& 32.89 & {\tiny $\pm$\hphantom{1}0.61}
& 50.84 & {\tiny $\pm$\hphantom{1}0.47}
& 83.98 & {\tiny $\pm$\hphantom{1}0.67} &
& 22.33 & {\tiny $\pm$\hphantom{1}0.92}
& 45.21 & {\tiny $\pm$\hphantom{1}1.05}
& 82.93 & {\tiny $\pm$\hphantom{1}0.25} \\

ViterbiPlanNet~\cite{vitrebiplannet}
& \cellcolor{tabsecond}33.99 & {\cellcolor{tabsecond}\tiny $\pm$\hphantom{1}0.23}
& 50.87 & {\tiny $\pm$\hphantom{1}0.17}
& 83.88 & {\tiny $\pm$\hphantom{1}0.31} &
& \cellcolor{tabsecond} 23.92 & {\cellcolor{tabsecond}\tiny $\pm$\hphantom{1}0.29}
& 45.63 & {\tiny $\pm$\hphantom{1}0.55}
& 82.56 & {\tiny $\pm$\hphantom{1}0.44} \\

\midrule
\methodname{}
& \cellcolor{tabfirst} 34.11 & {\cellcolor{tabfirst}\tiny $\pm$\hphantom{1}0.24} 
& \cellcolor{tabfirst} 51.18 & {\cellcolor{tabfirst}\tiny $\pm$\hphantom{1}0.22} 
& \cellcolor{tabfirst} 84.70 & {\cellcolor{tabfirst}\tiny $\pm$\hphantom{1}0.47} & \cellcolor{tabfirst}
& \cellcolor{tabfirst} 24.25 & {\cellcolor{tabfirst}\tiny $\pm$\hphantom{1}0.34} 
& \cellcolor{tabfirst} 46.13 & {\cellcolor{tabfirst}\tiny $\pm$\hphantom{1}0.63} 
& \cellcolor{tabsecond} 83.24 & {\cellcolor{tabsecond}\tiny $\pm$\hphantom{1}0.26} \\
\bottomrule
\end{tabularx}%
\label{tab:coin}
\end{table*}

\subsubsection{Implementation Details.} We implement our predictor model as a four-layer transformer \cite{transformer} with \num{6} attention heads, and a hidden dimension of \num{384}. We train the predictor using the AdamW optimizer \cite{adamw} and a learning rate of 5$\times$10\textsuperscript{-4}. We sample \num{50} negative samples per positive training sequence with $r=\text{\num{0.8}}$. The minimum $\tau_{\text{min}}$ and maximum margin $\tau_{\text{max}}$ are set to \num{0.01} and \num{0.1}, respectively. For obtaining the visual representations of the initial state and the goal state, we follow existing work \cite{ddn,kepp,schema,plate,pdpp,planllm,mtid} and use a frozen S3D~\cite{endtoend} encoder pre-trained on HowTo100M~\cite{howto100m}. Textual action sequences are encoded using the pre-trained text encoder from CLIP-ViT-B \cite{clip-orig} and kept frozen during training.

\subsection{Comparison to the State of The Art} \label{sec:mainresults}

We compare \methodname{} to recent state-of-the-art approaches for procedure planning in instructional videos. \Cref{tab:crosstask} presents the results on CrossTask for a planning horizon of $\rm T=\text{\num{3}}$ and $\rm T=\text{\num{4}}$. \methodname{} outperforms the recent state of the art on all procedure planning accuracy metrics and both planning horizons. Notably, \methodname{} outperforms recent approaches that utilize large language models (LLMs). Without relying on LLMs, \methodname{}, for example, outperforms the LLM-based approach SCHEMA~\cite{schema} by \SI{2.38}{\%} in SR for $\rm T=\text{\num{3}}$. In comparison to the recent state-of-the-art ViterbiPlanNet \cite{vitrebiplannet}, \methodname{} improves success rate by \SI{1.17}{\%} and \SI{0.12}{\%} for $\rm T=\text{\num{3}}$ and $\rm T=\text{\num{4}}$, respectively.

In \cref{tab:coin}, we report the results on the COIN dataset for a planning horizon $\rm T=\text{\num{3}}$ and $\rm T=\text{\num{4}}$. \methodname{} achieves the best mean SR, mAcc, and mIoU at $\rm T=\text{\num{3}}$, and the best SR and mAcc at $\rm T=\text{\num{4}}$. Improvements over the recent state-of-the-art ViterbiPlanNet~\cite{vitrebiplannet} are smaller on COIN than on CrossTask (\SI{0.12}{\%} and \SI{0.33}{\%} SR at $\rm T=\text{\num{3}}$ and $\rm T=\text{\num{4}}$, respectively). We attribute the smaller benefits to COIN's much wider task distribution (\num{180} tasks \emph{vs.}\ \num{18} on CrossTask), which reduces the number of training videos available per task and thereby weakens the intra-task hard-negative signal that drives our contrastive training. Even so, \methodname{} achieves state-of-the-art accuracy on COIN without significant additional training data, confirming the trend observed on CrossTask.

\begin{table}[t]
\centering
\scriptsize
\renewcommand{\arraystretch}{0.975}
\caption{\textbf{\methodname{} training analysis.} Component-wise analysis of \methodname{} on CrossTask val.\ and report SR, mACC \& mIoU (all in \%,\,$\uparrow$). We add components to the previous row until achieving \methodname{}. For reference, we include the recent state-of-the-art ViterbiPlanNet. Best results in red \colorindicator{tabfirst}; second-best results in orange \colorindicator{tabsecond}.}
\setlength{\tabcolsep}{4.75pt}
\begin{tabularx}{\linewidth}{@{}l S[table-format=2.2]S[table-format=2.2]S[table-format=2.2] l S[table-format=2.2]S[table-format=2.2]S[table-format=2.2]@{}}
\toprule
\raisebox{-3.0pt}[0pt][0pt]{\multirow{2}{*}{\textbf{Configuration}}}
& \multicolumn{3}{c}{\textbf{T=3}}
& \hphantom{4} & \multicolumn{3}{c}{\textbf{T=4}} \\
\cmidrule(lr){2-4} \cmidrule(lr){6-8}
& {SR\,$\uparrow$} & {mAcc\,$\uparrow$} & {mIoU\,$\uparrow$} &
& {SR\,$\uparrow$} & {mAcc\,$\uparrow$} & {mIoU\,$\uparrow$} \\
\midrule
ViterbiPlanNet~\cite{vitrebiplannet}
& 38.45 & 63.07 & 83.89 &
& 24.64 & 57.00 & 81.18 \\
\midrule
Baseline
& 13.65 & 47.82 & 69.53 &
& 9.24 & 35.44 & 67.76 \\
\ \ + {triplet contrastive loss}
& 37.58 & 63.21 & 83.67 &
& 23.54 & 56.18 & 80.91 \\
\ \ + {adaptive margin}
& \cellcolor{tabsecond} 38.74 & \cellcolor{tabsecond} 63.79 & \cellcolor{tabsecond} 84.02 & \cellcolor{tabsecond}
& \cellcolor{tabsecond} 24.23 & \cellcolor{tabsecond} 57.01 & \cellcolor{tabsecond} 81.22 \\
\ \ + {auxiliary regualrization (\methodname{})}
& \cellcolor{tabfirst} 39.62 & \cellcolor{tabfirst} 64.12 & \cellcolor{tabfirst} 84.29 & \cellcolor{tabfirst}
& \cellcolor{tabfirst} 24.76 & \cellcolor{tabfirst} 57.53 & \cellcolor{tabfirst} 81.58 \\
\bottomrule
\end{tabularx}
\label{tab:ablation}
\end{table}

\begin{table}[t]
\centering
\scriptsize
\renewcommand{\arraystretch}{0.975}
\caption{\textbf{Text encoder analysis.} We analyze the choice of text encoder for \methodname{} on CrossTask
val. and report SR, mACC, and mIoU (all in \%,\,$\uparrow$). Best results are highlighted in red \colorindicator{tabfirst} and second-best results in orange \colorindicator{tabsecond}.}
\setlength{\tabcolsep}{9.15pt}
\begin{tabularx}{\linewidth}{@{}lS[table-format=2.2]S[table-format=2.2]S[table-format=2.2] l S[table-format=2.2]S[table-format=2.2]S[table-format=2.2]@{}}
\toprule
\raisebox{-3.0pt}[0pt][0pt]{\multirow{2}{*}{\textbf{Text Encoder}}}
& \multicolumn{3}{c}{\textbf{T=3}}
& \hphantom{4} & \multicolumn{3}{c}{\textbf{T=4}} \\
\cmidrule(lr){2-4} \cmidrule(lr){6-8}
& {SR\,$\uparrow$} & {mAcc\,$\uparrow$} & {mIoU\,$\uparrow$} &
& {SR\,$\uparrow$} & {mAcc\,$\uparrow$} & {mIoU\,$\uparrow$} \\
\midrule
Random embedding & 35.10 & 61.20 & 82.50 &  & 21.30 & 54.10 & 79.80 \\
Flan-T5-base~\cite{flant5}           & \cellcolor{tabsecond} 37.80 & \cellcolor{tabsecond} 63.00 & \cellcolor{tabsecond} 83.60 & \cellcolor{tabsecond} & \cellcolor{tabsecond} 23.10 & \cellcolor{tabsecond} 56.20 & \cellcolor{tabsecond} 80.90 \\
CLIP~\cite{clip-orig}            & \cellcolor{tabfirst} 39.62 & \cellcolor{tabfirst} 64.12 & \cellcolor{tabfirst} 84.29 & \cellcolor{tabfirst} & \cellcolor{tabfirst} 24.76 & \cellcolor{tabfirst} 57.53 & \cellcolor{tabfirst} 81.58 \\
\bottomrule
\end{tabularx}%
\label{tab:text_encoder_ablation}
\end{table}

\subsection{Analyzing \methodname{}} \label{sec:analysis}
\subsubsection{Learning Objective.} Contrastive learning is the core component of \methodname{}, as it explicitly enforces a discriminative energy field in which the predicted goal is encoded near $x_g$ for correct sequences and far from $x_g$ for incorrect ones. In \cref{tab:ablation}, we report the contribution of each core component on CrossTask. The baseline replaces our contrastive objective with a plain $\mathcal{L}_2$ regression to $x_g$ and severely underperforms w.r.t.\ ViterbiPlanNet. Adding the triplet contrastive loss yields a $+\num{23.93}$ jump in SR for $\rm T=\text{\num{3}}$. The adaptive margin contributes an additional $+\SI{1.16}{\%}$ in SR by allowing hard negatives to receive a smaller separation than easy ones. Our auxiliary action-reconstruction loss further increases SR by $+\SI{0.88}{\%}$, encouraging informative auxiliary latents.

\subsubsection{Text Encoder Analysis.} We compare three options for embedding the action labels fed to $P_{\theta}$: a randomly initialized embedding trained from scratch, frozen Flan-T5-base~\cite{flant5}, and frozen CLIP (text encoder only)~\cite{clip-orig}. \Cref{tab:text_encoder_ablation} reports the results on CrossTask for $\rm T=\text{\num{3}}$ and $\rm T=\text{\num{4}}$ using the different options. Both pre-trained encoders outperform the random embedding across all metrics, indicating that pre-training provides a stronger foundation than what can be learned from the procedure-planning supervision alone. CLIP consistently outperforms Flan-T5-base despite being the smaller model. We attribute this to CLIP's multimodal contrastive pre-training, which aligns text representations with visual concepts and is well-aligned with our planning task.

\begin{figure}[t]
    \centering
    \pgfdeclarelayer{foreground}
\pgfdeclarelayer{background}
\pgfsetlayers{background,main,foreground}
\begin{tikzpicture}[clip, every node/.style={font=\sffamily\fontsize{7}{2}}, >={Stealth[inset=0pt,length=3.5pt,angle'=45]}]
  \tikzset{every picture/.style={/utils/exec={\sffamily\fontsize{7}{2}}}}
  \tikzset{every major tick/.append style={line width=.5pt, major tick length=3.5pt, gray!85}}

  \pgfkeys{/pgfplots/CommonAxisStyle/.style={
      enlargelimits=false,
      clip=true,
      height=2.9cm,
      width=0.575\textwidth,
      grid=both,
      xtick pos=bottom,
      ylabel shift=-2.0pt,
      ytick pos=left,
      grid style={line width=.1pt, draw=white, dash pattern=on 1pt off 1pt},
      major grid style={line width=.2pt,draw=gray!65},
      minor tick num=1,
  }}

  \begin{groupplot}[group style={group size=2 by 1, horizontal sep=5mm}, CommonAxisStyle]

    \nextgroupplot[
      title={\sffamily\fontsize{7}{2}${\rm T}=\text{3}$; $r=\text{0.8}$},
      title style={yshift=-5pt},
      ylabel=SR $\uparrow$,
      xlabel={Number of negative samples $N$},
      xlabel shift=-2.0pt,
      ymin=33, ymax=41,
      xmode=log,
      ytick={34, 36, 38, 40},
      yticklabels={34, 36, 38, 40},
      xmin=8, xmax=350,
      xtick={10, 25, 50, 100, 250},
      xticklabels={10, 25, 50, 100, 250},
      ticklabel style = {font=\fontsize{7}{2}},
      ]

      \addplot[color=tud1c, dash pattern=on 4pt off 2.45pt, thick, mark=diamond*, mark options={solid}, mark size=1.5pt,
        error bars/.cd, y dir=both, y explicit] table[x=N,y=SR] {negativesamples.dat};

      \addplot[color=tud9c, dash pattern=on 4pt off 2.45pt, thick, mark=diamond*, mark options={solid}, mark size=2.25pt,
        error bars/.cd, y dir=both, y explicit] table[x=N,y=SR] {negativesamplesbest.dat};

    \nextgroupplot[
      title={${\rm T}=\text{3}$; $N=\text{50}$},
      title style={yshift=-5pt},
      ylabel={},
      xlabel={Hard/easy negative sample ratio $r$},
      xlabel shift=-2.0pt,
      ymin=33, ymax=41,
      xmin=-0.1, xmax=1.1,
      ytick={34, 36, 38, 40},
      yticklabels={34, 36, 38, 40},
      xtick={0, 0.25, 0.5, 0.8, 1},
      xticklabels={0, 0.25, 0.5, 0.8, 1},
    ]
        \addplot[color=tud7c, dash pattern=on 4pt off 2.45pt, thick, mark=diamond*, mark options={solid}, mark size=1.5pt,
        error bars/.cd, y dir=both, y explicit] table[x=r,y=SR] {ratio.dat};
        \addplot[color=tud9c, dash pattern=on 4pt off 2.45pt, thick, mark=diamond*, mark options={solid}, mark size=2.25pt,
        error bars/.cd, y dir=both, y explicit] table[x=r,y=SR] {ratiobest.dat};

        \draw[<-, thick] (axis cs:0, 33.65) -- (axis cs:0.1, 33.65) node[right, right=0pt, fill=white, fill opacity=0.65, text opacity=1.0, font=\sffamily\tiny, inner sep=0.3pt, anchor=west] {Only easy neg.};

        \draw[<-, thick] (axis cs:1.0, 33.65) -- (axis cs:0.9, 33.65) node[right, right=0pt, fill=white, fill opacity=0.65, text opacity=1.0, font=\sffamily\tiny, inner sep=0.3pt, anchor=east] {Only hard neg.};
    
  \end{groupplot}
\end{tikzpicture}%
    \vspace{-1.0cm}
    \caption{\textbf{Negative sampling analysis}. We analyze the impact of the number of negative samples $N$ \emph{(left)} and the hard/easy negative sample ratio $r$ \emph{(right)} on CrossTask with $\rm T=\text{\num{3}}$ using SR (in \%, $\uparrow$). For $r=\text{\num{0}}$, only easy negatives are used. \emph{Vice versa}, $r=\text{\num{1}}$ generates only hard negatives. We indicate our default values in red \colorindicator{tud9c}.}
    \vspace{5pt}
    \label{fig:ne-sample-ablation}
\end{figure}
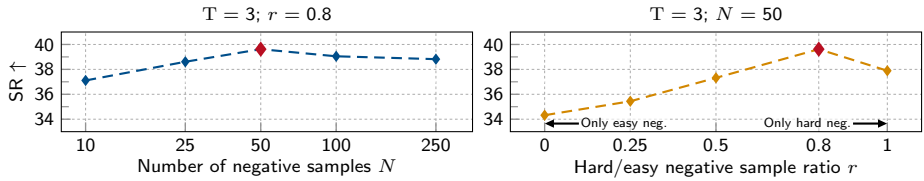
\begin{table}[t]
    \centering
    \begin{minipage}[t]{0.315\linewidth}
            \centering
            \scriptsize
            \renewcommand{\arraystretch}{0.975}
            \caption{\textbf{Task classifier results.} We report the accuracy (in \%, $\uparrow$) of our task classifier on the high-level class tasks of CrossTask \cite{crosstask} as well as COIN~\cite{coin} for two different planning horizons.}
            \vspace{2.5pt}
            \setlength{\tabcolsep}{6.0pt}
            \label{tab:task-classification}
            \begin{tabularx}{\textwidth}{@{}XS[table-format=2.2]S[table-format=2.2]}
                \toprule
                \textbf{Dataset} & {\textbf{T=3}} & {\textbf{T=4}} \\
                \midrule
                CrossTask & 92.43 & 92.98 \\
                COIN      & 79.42 & 79.42 \\
                \bottomrule
            \end{tabularx}
    \end{minipage}%
    \hfill%
    \begin{minipage}[t]{0.66\linewidth}
        \centering
\scriptsize
\renewcommand{\arraystretch}{0.975}
\caption{\textbf{Oracle experiment.} We report oracle results by replacing the task classifier prediction used for inference with the ground truth task on CrossTask and COIN, using SR, mACC \& mIoU (all in \%,\,$\uparrow$). For reference, we also report \methodname{} (\emph{i.e.}, w/ task classifier) in \colorindicator{gray!70}.}
\setlength{\tabcolsep}{1.5pt}
\begin{tabularx}{\linewidth}{@{}ll S[table-format=2.2]S[table-format=2.2]S[table-format=2.2] l S[table-format=2.2]S[table-format=2.2]S[table-format=2.2]@{}}
\toprule
\raisebox{-3.0pt}[0pt][0pt]{\multirow{2}{*}{\textbf{Datasets\vphantom{g}}}}
& \raisebox{-3.0pt}[0pt][0pt]{\multirow{2}{*}{\textbf{Setting}}}
& \multicolumn{3}{c}{\textbf{T=3}}
& & \multicolumn{3}{c}{\textbf{T=4}} \\
\cmidrule(lr){3-5} \cmidrule(lr){7-9}
& & {SR\,$\uparrow$} & {mAcc\,$\uparrow$} & {mIoU\,$\uparrow$} &
& {SR\,$\uparrow$} & {mAcc\,$\uparrow$} & {mIoU\,$\uparrow$} \\
\midrule
\multirow{2}{*}{CrossTask}
& \methodname{} &\color{gray!70} 39.62 &\color{gray!70} 64.12 &\color{gray!70} 84.29 & &\color{gray!70} 24.76 &\color{gray!70} 57.53 &\color{gray!70} 81.58 \\
& Oracle & 41.32 & 66.42 & 88.24 &  & 25.52 & 58.85 & 85.02 \\
\midrule
\multirow{2}{*}{COIN}
& \methodname{} &\color{gray!70} 34.11 &\color{gray!70} 51.18 &\color{gray!70} 84.70 & &\color{gray!70} 24.25 &\color{gray!70} 46.13 &\color{gray!70} 83.24 \\
& Oracle & 38.33 & 59.62 & 98.15 & & 29.51 & 55.85 & 97.95\\
\bottomrule
\end{tabularx}
\label{tab:oracle_comparison}

    \end{minipage}%
\end{table}

\subsubsection{Negative-Sequence Selection.} We analyze the two hyperparameters of our negative-sampling strategy: the number of negatives $N$ and the hard/easy negative sample ratio $r$. \Cref{fig:ne-sample-ablation} reports the success rate on CrossTask with $\rm T=\text{\num{3}}$. Success rate is low for $N=\text{\num{10}}$ due to insufficient contrastive supervision and peaks at $N=\text{\num{50}}$. Further increasing $N$ does not lead to improvements; accuracy saturates. The hard-negative ratio interpolates between only easy ($r=\text{\num{0}}$, inter-task) and only hard ($r=\text{\num{1}}$, intra-task) negatives. Performance peaks at $r=\text{\num{0.8}}$, indicating that a mixture biased towards hard intra-task negatives provides the strongest supervision while retaining inter-task separation.

\subsubsection{Task Classifier Analysis.} 
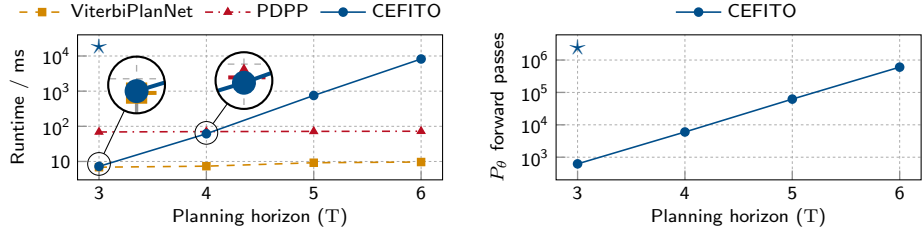
\begin{figure}[t]
    \centering
    \pgfdeclareplotmark{mystar}{
    \node[] {\large $\star$};
}

\begin{tikzpicture}[
    every node/.style={font=\sffamily\fontsize{7}{2}},
    spy using outlines={circle,magnification=2.5,size=0.75cm,connect spies,very thick, every spy in node/.append style={rotate=0, thick, fill=white}},
]
\tikzset{every picture/.style={/utils/exec={\sffamily\fontsize{7}{2}}}}
  \tikzset{every major tick/.append style={line width=.5pt, major tick length=3.5pt, gray!85}}

\pgfkeys{/pgfplots/CommonAxisStyle/.style={
    enlargelimits=false,
    clip=true,
    height=3.5cm,
    width=0.525\linewidth,
    grid=both,
    xtick pos=bottom,
    ytick pos=left,
    grid style={line width=.1pt, draw=white, dash pattern=on 1pt off 1pt},
      major grid style={line width=.2pt,draw=gray!65},
    minor tick num=0,
}}

\begin{groupplot}[group style={group size=2 by 1, horizontal sep=15mm}, CommonAxisStyle]
    \nextgroupplot[
      CommonAxisStyle,
    xlabel={Planning horizon ({$\rm T$})},
    ylabel={Runtime / ms},
    xmin=2.8,
    xmax=6.2,
    xtick={3,4,5,6},
    ymode=log,
    log basis y={10},
    xlabel shift=-2.0pt,
    ymin=3,
    ymax=40000,
    ytick={1,10,100,1000,10000},
    yticklabels={1,10,10\textsuperscript{2},10\textsuperscript{3},10\textsuperscript{4}},
    xticklabels={3,4,5,6},
    legend style={
        at={(0.425,1.025)},
        anchor=south,
        legend columns=3,
        draw=none,
        fill=none,
        font=\sffamily\fontsize{7}{2}
    },
    ]

\addplot[
    color=tud7c,
    semithick,
    dashed,
    mark=square*,
    mark size=1.3pt,
    mark options={solid},
] table[x=T,y=ViterbiPlanNet]{inference.dat};
\addlegendentry{ViterbiPlanNet\;};

\addplot[
    color=tud9c,
    semithick,
    dashdotted,
    mark=triangle*,
    mark size=1.5pt,
    mark options={solid},
] table[x=T,y=PDPP]{inference.dat};
\addlegendentry{PDPP\;};

\addplot[
    color=tud1c,
    semithick,
    mark=*,
    mark size=1.5pt,
    mark options={solid},
] table[x=T,y=CEFITO]{inference.dat};
\addlegendentry{CEFITO};

\addplot[
    color=tud1c,
    semithick,
    mark=mystar,
    mark size=1.5pt,
    mark options={solid},
] table[x=T,y=CEFITOFull]{inference.dat};

\coordinate (spypoint1) at (axis cs:3,8.5);
\coordinate (spypos1) at (axis cs:3.35,1500);
\coordinate (spypoint2) at (axis cs:4,65);
\coordinate (spypos2) at (axis cs:4.35,2000);

\nextgroupplot[
      CommonAxisStyle,
    xlabel={Planning horizon ({$\rm T$})},
    ylabel={$P_{\theta}$ forward passes},
    xmin=2.8,
    xmax=6.2,
    xtick={3,4,5,6},
    ymode=log,
    log basis y={10},
    xlabel shift=-2.0pt,
    ymin=200,
    ymax=6000000,
    ytick={1e3,1e4,1e5,1e6},
    yticklabels={10\textsuperscript{3},10\textsuperscript{4},10\textsuperscript{5},10\textsuperscript{6}},
    xticklabels={3,4,5,6},
    legend style={
        at={(0.5,1.025)},
        anchor=south,
        legend columns=1,
        draw=none,
        fill=none,
        font=\sffamily\fontsize{7}{2}
    },
    ]
\addplot[
    color=tud1c,
    semithick,
    mark=*,
    mark size=1.5pt,
    mark options={solid},
] table[x=T,y=n_candidates]{inference.dat};
\addlegendentry{CEFITO};

\addplot[
    color=tud1c,
    semithick,
    mark=mystar,
    mark size=1.5pt,
    mark options={solid},
] table[x=T,y=n_candidatesfull]{inference.dat};

\end{groupplot}

\spy[black] on (spypoint1) in node at (spypos1);
\spy[black] on (spypoint2) in node at (spypos2);
\end{tikzpicture}
    \vspace{-1.0cm}
    \caption{\textbf{Inference runtime results}. \emph{Left:} We report inference runtime (in ms, $\downarrow$) over different planning horizons for inferring a single action sequence on the CrossTask dataset \cite{crosstask}. All runtimes are reported using the same hardware (\emph{single} A100 \SI{80}{GB} GPU). \emph{Right:} We report the average number of forward passes through our predictor $P_{\theta}$ required for inference. While \methodname{} (in \colorindicator{tud1c}) provides a comparable inference runtime for $\rm T\leq 4$ to diffusion (PDPP \cite{pdpp} in \colorindicator{tud9c}) and feed-forward (ViterbiPlanNet \cite{vitrebiplannet} in \colorindicator{tud7c}) approaches, the runtime is significantly worse for $\rm T\geq 5$. For reference, we also report \methodname{} without constraining the search space, indicated using \textcolor{tud1c}{\large $\star$}.}
    \label{fig:inference-time-ablation}
\end{figure}

During inference, we use the prediction of our task classifier to effectively constrain the search space and reduce runtime (\emph{cf.} \cref{fig:inference-time-ablation}). However, an incorrect task prediction always results in a wrong action sequence. To analyze this, we report the accuracy of our classifier in \cref{tab:task-classification}. On CrossTask, we achieve an accuracy above \SI{90}{\%}, introducing only a minor error. On COIN, our classifier introduces a more significant error. In about \SI{20}{\%} of the validation samples, our task classifier is the cause of a wrong action sequence prediction.

We demonstrated that our task classifier can make initial errors, causing inference to fail (\emph{cf.} \cref{tab:task-classification}). To analyze the impact of task classifier errors on the downstream accuracy, we perform an oracle experiment in \cref{tab:oracle_comparison}. In particular, we replace the classifier's high-level task prediction with the ground truth and then run inference. This provides an upper bound on the downstream accuracy of \methodname{}. While using the ground-truth class for inference (oracle setting) yields consistent improvements on both datasets, the improvements on COIN are more significant. This demonstrates that an improved task classification accuracy can directly improve downstream planning.

\subsubsection{Limitations \& Failed Experiments.} \methodname{} achieves state-of-the-art accuracy on CrossTask (\emph{cf.} \cref{tab:crosstask}) and COIN (\emph{cf.} \cref{tab:coin}), demonstrating that learning an expressive action-condition energy field and performing task-constrained optimization is feasible and effective. While we restrict the set of possible actions using a learned task constraint, significantly reducing runtime (\emph{cf.} \cref{fig:inference-time-ablation}), search complexity still grows exponentially with the planning horizon $\rm T$. We demonstrate this empirically in \cref{fig:inference-time-ablation}. For shorter sequences, inference runtime is manageable and comparable to existing methods (\emph{cf.} \cref{fig:inference-time-ablation} \emph{(left)}). For larger sequences $\rm T \geq \text{\num{5}}$, the runtime of \methodname{} significantly increases as more predictor forward passes need to be performed (\emph{cf.} \cref{fig:inference-time-ablation} \emph{(right)}). Concretely, for a planning horizon of \num{5}, \methodname{} is about one order of magnitude slower than PDPP. For $\rm T =\text{\num{6}}$ \methodname{} is even two orders of magnitude slower. To overcome this limitation, approximate optimization strategies, such as beam search \cite{Lowerre:1976:BES}, offer a potential avenue to reduce inference runtime. While we optimize over a set of discrete actions, adapting gradient-based inference \cite{Belanger:2016:ENN} to this setting could provide an additional avenue for improving inference-time optimization runtime.

\begin{table*}[t]
\centering
\renewcommand{\arraystretch}{0.975}
\caption{\textbf{Results on NIV \cite{niv}.} We compare \methodname{} with the state-of-the-art on NIV val., using different planning horizons, and report SR, mACC, and mIoU (all in \%, $\uparrow$). Best results are highlighted in red \colorindicator{tabfirst} and second-best results in orange \colorindicator{tabsecond}.}
\setlength{\tabcolsep}{1.34pt}
\scriptsize
\sisetup{table-number-alignment=center}
\begin{tabularx}{\linewidth}{@{}l S[table-format=2.2]lS[table-format=2.2]lS[table-format=2.2]l l S[table-format=2.2]lS[table-format=2.2]lS[table-format=2.2]l@{}}
\toprule
\raisebox{-3.0pt}[0pt][0pt]{\multirow{2}{*}{\textbf{Method}}} 
& \multicolumn{6}{c}{\textbf{T=3}} 
& \hphantom{4} & \multicolumn{6}{c}{\textbf{T=4}} \\
\cmidrule(lr){2-7} \cmidrule(lr){9-14}
& \multicolumn{2}{c}{{SR\,$\uparrow$}} & \multicolumn{2}{c}{mAcc\,$\uparrow$} & \multicolumn{2}{c}{mIoU\,$\uparrow$} 
& & \multicolumn{2}{c}{SR\,$\uparrow$} & \multicolumn{2}{c}{mAcc\,$\uparrow$} & \multicolumn{2}{c}{mIoU\,$\uparrow$} \\
\midrule

Qwen2.5-VL-32B~\cite{qwen-2.5-vl}
& 7.41 & & 27.65 & & 59.73 & &
& 5.26 & & 28.84 & & 60.21 & \\

Qwen2.5-32B~\cite{qwen-2.5}
& 24.07 & & 43.46 & & 71.88 & &  
& 23.25 & & 41.89 & & 73.91 & \\

Gemini 2.5 Pro~\cite{gemini-2.5}
& 24.07 & & 43.46 & & 71.86 & &   
& 22.37 & & 40.35 & & 73.05 & \\

Qwen3-30B~\cite{qwen3}
& 24.81 & & 42.84 & & 70.80 & &    
& 22.37 & & 41.23 & & 73.90 & \\

Qwen3-30B + PKG~\cite{vitrebiplannet}  
& 25.19 & & 43.95 & & 71.98 & &       
& 21.93 & & 41.67 & & 74.43 & \\

PKG beam search~\cite{vitrebiplannet}
& 24.96 & {\tiny$\pm$\num{1.93}} 
& 43.46 & {\tiny$\pm$\num{2.42}} 
& 72.18 & {\tiny$\pm$\num{0.55}} &
& 21.23 & {\tiny$\pm$\num{0.96}} 
& 40.86 & {\tiny$\pm$\num{0.83}} 
& 72.69 & {\tiny$\pm$\num{0.75}} \\

PDPP~\cite{pdpp}
& 26.52 & {\tiny$\pm$\num{1.56}} 
& 45.58 & {\tiny$\pm$\num{1.85}} 
& \cellcolor{tabfirst} 74.89 & {\cellcolor{tabfirst} \tiny$\pm$\num{0.85}} &
& 21.40 & {\tiny$\pm$\num{0.53}} 
& 40.20 & {\tiny$\pm$\num{2.00}} 
& 72.82 & {\tiny$\pm$\num{1.84}} \\

KEPP~\cite{kepp}
& 27.56 & {\tiny$\pm$\num{1.48}} 
& \cellcolor{tabsecond} 45.93 & \cellcolor{tabsecond}{\tiny$\pm$\num{2.37}} 
& \cellcolor{tabsecond} 74.36 & \cellcolor{tabsecond} {\tiny$\pm$\num{0.97}} &
& 22.54 & {\tiny$\pm$\num{1.93}} 
& \cellcolor{tabsecond} 42.46 & \cellcolor{tabsecond}{\tiny$\pm$\num{1.49}} 
& 73.11 & {\tiny$\pm$\num{0.94}} \\

PlanLLM~\cite{planllm}
& \cellcolor{tabsecond} 30.00 & \cellcolor{tabsecond}{\tiny$\pm$\num{1.41}} 
& 44.35 & {\tiny$\pm$\num{2.52}} 
& 73.60 & {\tiny$\pm$\num{1.66}} &
& 23.42 & {\tiny$\pm$\num{1.40}} 
& 41.95 & {\tiny$\pm$\num{2.81}} 
& 72.32 & {\tiny$\pm$\num{0.91}} \\

SCHEMA~\cite{schema}
& 26.30 & {\tiny$\pm$\num{1.49}} 
& 42.77 & {\tiny$\pm$\num{2.12}} 
& 73.04 & {\tiny$\pm$\num{1.42}} &
& 24.39 & {\tiny$\pm$\num{1.84}} 
& 41.14 & {\tiny$\pm$\num{3.62}} 
& 73.13 & {\tiny$\pm$\num{1.97}} \\

ViterbiPlanNet~\cite{vitrebiplannet}
& \cellcolor{tabfirst} 32.37 & \cellcolor{tabfirst}{\tiny$\pm$\num{0.96}} 
& \cellcolor{tabfirst} 46.96 & \cellcolor{tabfirst}{\tiny$\pm$\num{1.75}} 
& 73.85 & {\tiny$\pm$\num{0.85}} &
& \cellcolor{tabfirst} 27.54 & \cellcolor{tabfirst}{\tiny$\pm$\num{0.70}} 
& \cellcolor{tabfirst} 45.55 & \cellcolor{tabfirst} {\tiny$\pm$\num{1.89}} 
& \cellcolor{tabfirst} 74.71 & \cellcolor{tabfirst}{\tiny$\pm$\num{1.19}} \\

\midrule
\methodname{} (Ours)
& 22.46 & {\tiny$\pm1.43$} 
& 40.37 & {\tiny$\pm2.21$} 
& 70.26 & {\tiny$\pm1.43$} &
& 19.67 & {\tiny$\pm1.76$} 
& 37.25 & {\tiny$\pm1.97$} 
& 70.75 & {\tiny$\pm1.39$} \\

\bottomrule
\end{tabularx}
\label{tab:niv}

\end{table*}

Beyond the strong results on CrossTask and COIN (\emph{cf.} Tabs.\ \ref{tab:crosstask} \& \ref{tab:coin}), we also present a failed experiment. In particular, we report results on NIV \cite{niv} in \cref{tab:niv}. \methodname{} yields a suboptimal accuracy. We root this in the fact that NIV is significantly smaller than both CrossTask and COIN. COIN comprises over \SI{11}{k} videos, NIV only contains \num{150}, about two orders of magnitude less. We suspect that contrastive learning collapses for a small number of training videos.

\section{Conclusion} \label{sec:conclusion}

We introduced \methodname{}, an inference-time optimization framework for procedure planning in instructional videos. By learning a predictor model that maps a candidate action sequence and the initial state to a predicted goal embedding, we can approach planning using task-constrained inference-time optimization. \methodname{} demonstrates state-of-the-art accuracy on CrossTask and COIN. Unlike current approaches, we plan using optimization at inference-time and do not rely on large language models.

\small{
{\subsubsection{Acknowledgments.} This project has received funding from the European Research Council (ERC) under the European Union’s Horizon 2020 research and innovation programme (grant agreement No.\ 866008). Additionally, this project is also funded by the Deutsche Forschungsgemeinschaft (DFG, German Research Foundation) under Germany\textquotesingle{}s Excellence Strategy (EXC-3066/1 ``The Adaptive Mind'', Project No.\ 533717223, EXC-3057/1 ``Reasonable Artificial Intelligence'', Project No.\ 533677015). Mohamed Afham \& Christoph Reich are supported by the Konrad Zuse School of Excellence in Learning and Intelligent Systems (\href{https://eliza.school}{ELIZA}) through the DAAD programme Konrad Zuse Schools of Excellence in Artificial Intelligence, sponsored by the Federal Ministry of Education and Research. This work was also supported by the ERC Advanced Grant SIMULACRON, the Georg Nemetschek Institute project AI4TWINNING, and the DFG project 4D-YouTube CR 250/26-1. %
}}

\bibliographystyle{splncs04}
\bibliography{egbib}

\clearpage
\appendix
\renewcommand{\thepage}{\roman{page}}
\setcounter{page}{1}
\renewcommand{\thefigure}{A.\arabic{figure}}
\setcounter{figure}{0}
\renewcommand{\thetable}{A.\arabic{table}} 
\setcounter{table}{0}
\renewcommand{\theequation}{A.\arabic{equation}} 
\setcounter{equation}{0}

\maketitlesupplementary
\thispagestyle{empty}
\label{app:supplement-master}
\FloatBarrier

\noindent{}In this supplement, we provide additional details on our auxiliary action sequence reconstruction (\emph{cf.} \cref{sec:auxiliary}), further implementation details (\emph{cf.} \cref{sec:implementation_details}), a summary of the core baseline methods we compare against (\emph{cf.} \cref{sec:baselines}), and qualitative results (\emph{cf.} \cref{sec:qualitative_results}).

\section{Auxiliary Action Sequence Reconstruction} \label{sec:auxiliary}

\begin{figure}[t]
    \centering
    \includegraphics[width=0.999\linewidth]{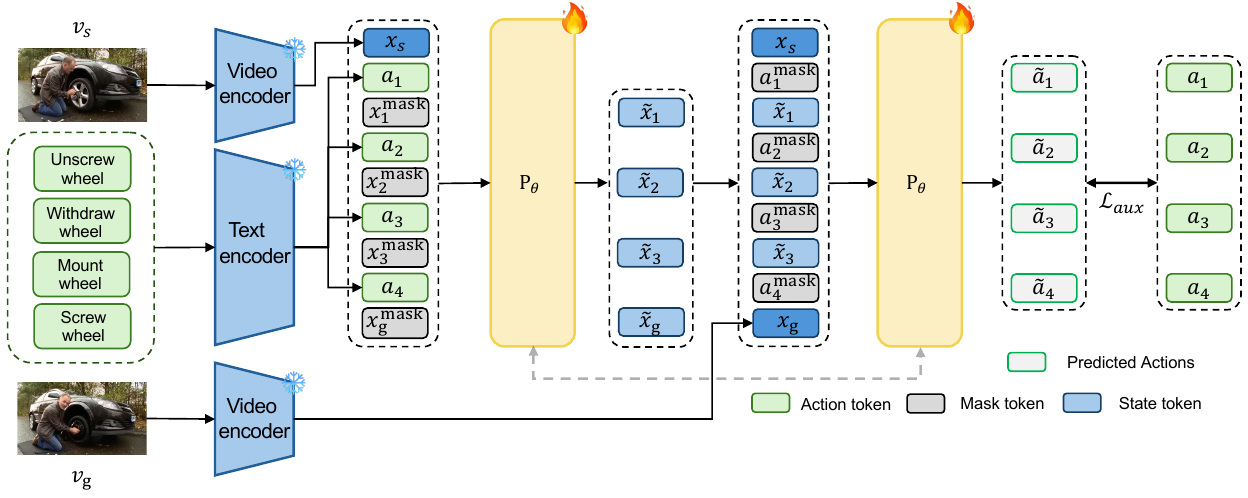}
    \caption{\textbf{Auxiliary action sequence reconstruction.} During training, the predictor $P_\theta$ is applied twice with shared weights: \emph{first}, given $x_s$, action tokens $a_{1:4}$, and masked state tokens $x_{1:4}^{\text{mask}}$, $P_\theta$ predicts the intermediate and goal latents $\tilde{x}_{1:3}, \tilde{x}_g$, respectivally. \emph{Second}, given $x_s$, $x_g$, $P_\theta$ predicts the masked action tokens. These are supervised by the ground-truth action sequence $a_{1:4}$ using the auxiliary loss $\mathcal{L}_{\text{aux}}$.}     
    \label{fig:l-aux}
\end{figure}

In addition to the contrastive loss $\mathcal{L}_{\text{c}}$, which only supervises the predicted goal state $\tilde{x}_{g}$, we employ an auxiliary action sequence reconstruction loss $\mathcal{L}_{\text{aux}}$ to obtain expressive intermediate representations. The initial and goal observations $v_s$ and $v_g$ are encoded by the video encoder into latent states $x_s$ and $x_g$, while the ground-truth action sequence is encoded by the text encoder into action tokens $a_{1:\rm T}$. The predictor $P_\theta$ is then applied in two passes to reconstruct the actions from the predicted intermediate representations. First pass (state prediction): $x_s$, the action tokens $a_{1:\rm T}$, and learnable mask tokens representing the unobserved intermediate and goal states ($\{x_{l}^{\text{mask}}\}_{t=1}^{\rm T-1}, x_{g}^{\text{mask}}$) are fed jointly into $P_\theta$, which produces the predicted latent states $\{\tilde{x}_{l}\}_{t=1}^{\rm T-1}, \tilde{x}_{g}$. Second pass (action reconstruction): the initial state token $x_s$, the predicted intermediate state latents $\{\tilde{x}_{l}\}_{t=1}^{\rm T-1}$, ground truth goal state token $x_g$, and learnable mask tokens representing the actions ($\{a_{k}^{\text{mask}}\}_{k=1}^{\rm T}$) are fed back into the same predictor (weight-shared) $P_\theta$, which this time predicts the actions $\tilde{a}_{1:\rm T}$. The predicted action tokens are mapped back to the action space, and the auxiliary loss $\mathcal{L}_{\text{aux}}$ penalizes the discrepancy between the reconstructed actions $\tilde{a}_{1:\rm T}$ and the ground-truth actions $a_{1:\rm T}$. The auxiliary action sequence reconstruction pipeline is visualized in \cref{fig:l-aux}.

\section{Implementation Details} \label{sec:implementation_details}
\subsubsection{Architecture.} 
We implement our predictor $P_\theta$ as a causal (masked) transformer~\cite{transformer}. We use four transformer blocks, each with six attention heads and a hidden dimension of \num{384}. In total, $P_\theta$ contains \SI{8.5}{M} learnable parameters. Into $P_\theta$, we feed the visual features of the initial $x_s$ and goal state $x_g$, as well as tokenized actions. We obtain visual features using the S3D~\cite{endtoend} visual encoder (pre-trained on HowTo100M~\cite{howto100m}) and a temporal window of \num{3}-frames surrounding $v_{s}$ and $v_{g}$, respectively. Before being fed into $P_\theta$, $x_s$ and $x_g$ are normalized along the feature dimension before being linearly projected to the hidden dimension of $P_\theta$. To feed textual actions into $P_\theta$, we tokenize and encode actions in natural language using the pre-trained text encoder from CLIP-ViT-B \cite{clip-orig}, before linearly projecting the resulting tokens to the hidden dimension of $P_\theta$. CLIP and the visual encoder are frozen; only the linear projections are learned. Our task classifier is implemented as a four-layer multilayer perceptron that takes in both $x_s$ and $x_g$.

\subsubsection{Training.} We train the predictor's weights $\theta$ using the AdamW optimizer~\cite{adamw}. AdamW uses a weight decay of \num{1e-3} and a learning rate of \num{5e-4}. \num{50} negative samples are generated per positive training sequence. We set the hard/easy negative sample ratio $r$ to \num{0.8}. The minimum $\tau_{\text{min}}$ margin is set to \num{0.01}. The maximum margin $\tau_{\text{max}}$ is set to \num{0.01}. We train $P_\theta$ for \num{200} epochs, using early stopping, following KEPP \cite{kepp}. The task classifier is trained using a learning rate of \num{1e-4} with no weight decay.

\section{Baselines} \label{sec:baselines}

\subsubsection{PDPP~\cite{pdpp}} approaches procedure planning as a distribution fitting problem under given observations, the initial and goal state. Learned using diffusion, PDPP can sample action-sequence predictions during inference. In particular, PDPP performs conditional diffusion and learns using action-sequence ground truth, and can express uncertainty in its prediction.

\subsubsection{KEPP~\cite{kepp}} extends PDPP by introducing a probabilistic procedure knowledge graph to the model's architecture. This knowledge graph is constructed using training data by iteratively adding and reweighing edges and serves as a retrieval
signal for diffusion-based inference.

\subsubsection{SCHEMA~\cite{schema}} demonstrates that state changes matter for procedure planning, and builds a structured state space by explicitly representing each step as a state change. A large language model (LLM) is used to generate language descriptions of these state changes. Then these descriptions are aligned with visual observations via cross-modal contrastive learning to track intermediate states.

\subsubsection{PlanLLM~\cite{planllm}} builds on SCHEMA's use of LLM-generated state descriptions, but instead of decoding into a fixed, closed set of action vectors, it lets the LLM generate free-form planning output. It adds an LLM-enhanced-planning module for flexible step decoding and a mutual-information-maximization module to link commonsense text with visual state

\subsubsection{ViterbiPlanNet~\cite{vitrebiplannet}} approaches procedure planning by explicitly incorporating procedural knowledge into the end-to-end training, different from KEPP. It introduces a differentiable Viterbi layer that embeds a procedural knowledge graph into Viterbi decoding, using smooth relaxations to enable end-to-end training.

\section{Qualitative Results} \label{sec:qualitative_results}

\begin{figure}[t]
    \centering
    \includegraphics[width=0.875\linewidth]{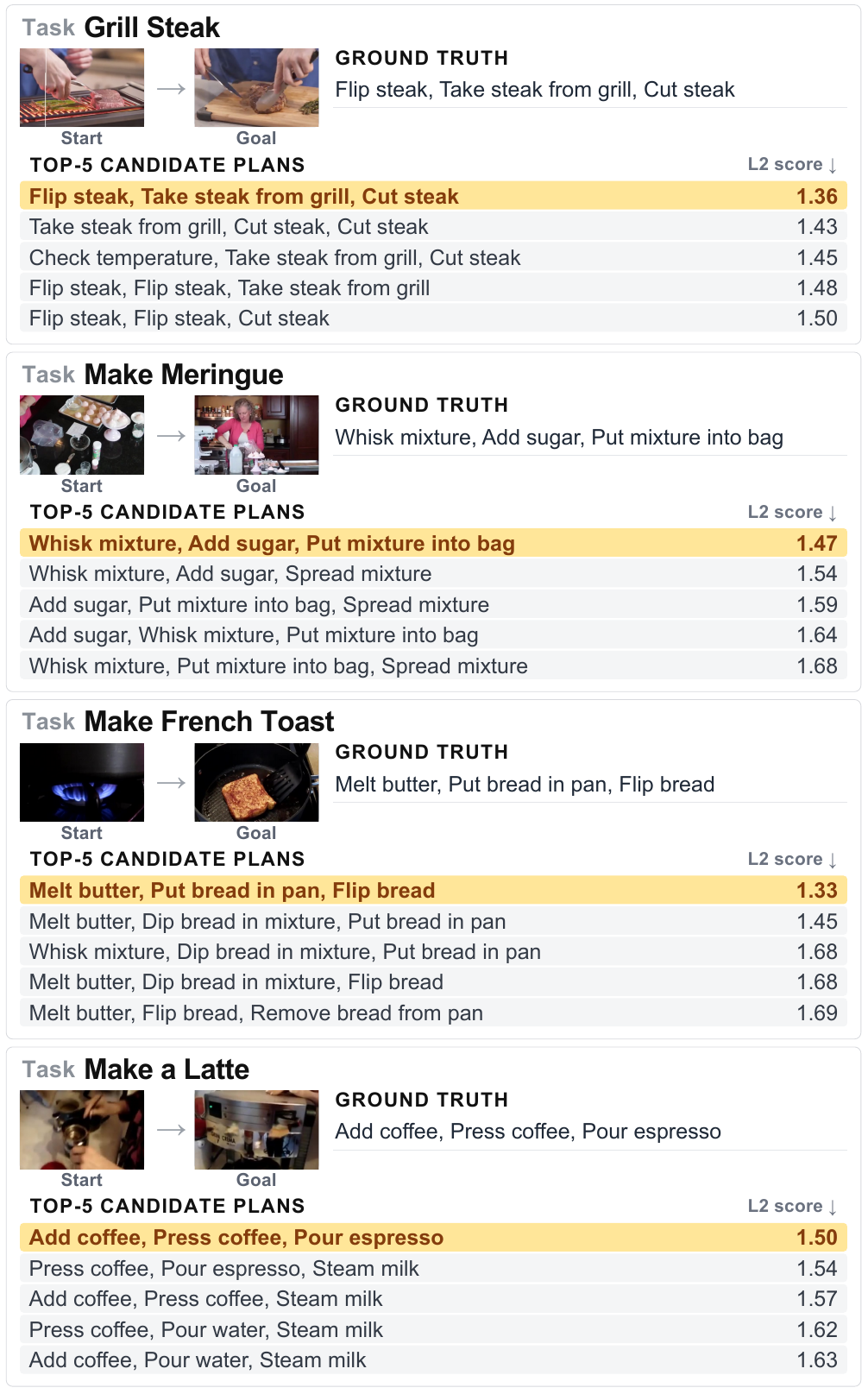}
    \caption{\textbf{Qualitative examples of \methodname{} on CrossTask.} Procedural video results where the top-ranked candidate plan (highlighted in \colorindicator{myyellow}) matches the ground-truth action sequence, retrieved from the top-5 candidates ranked by L2 distance to the goal embedding $x_{g}$.}   
    \label{fig:qualitative_success-1}
\end{figure}

\begin{figure}[t]
    \centering
    \includegraphics[width=0.875\linewidth]{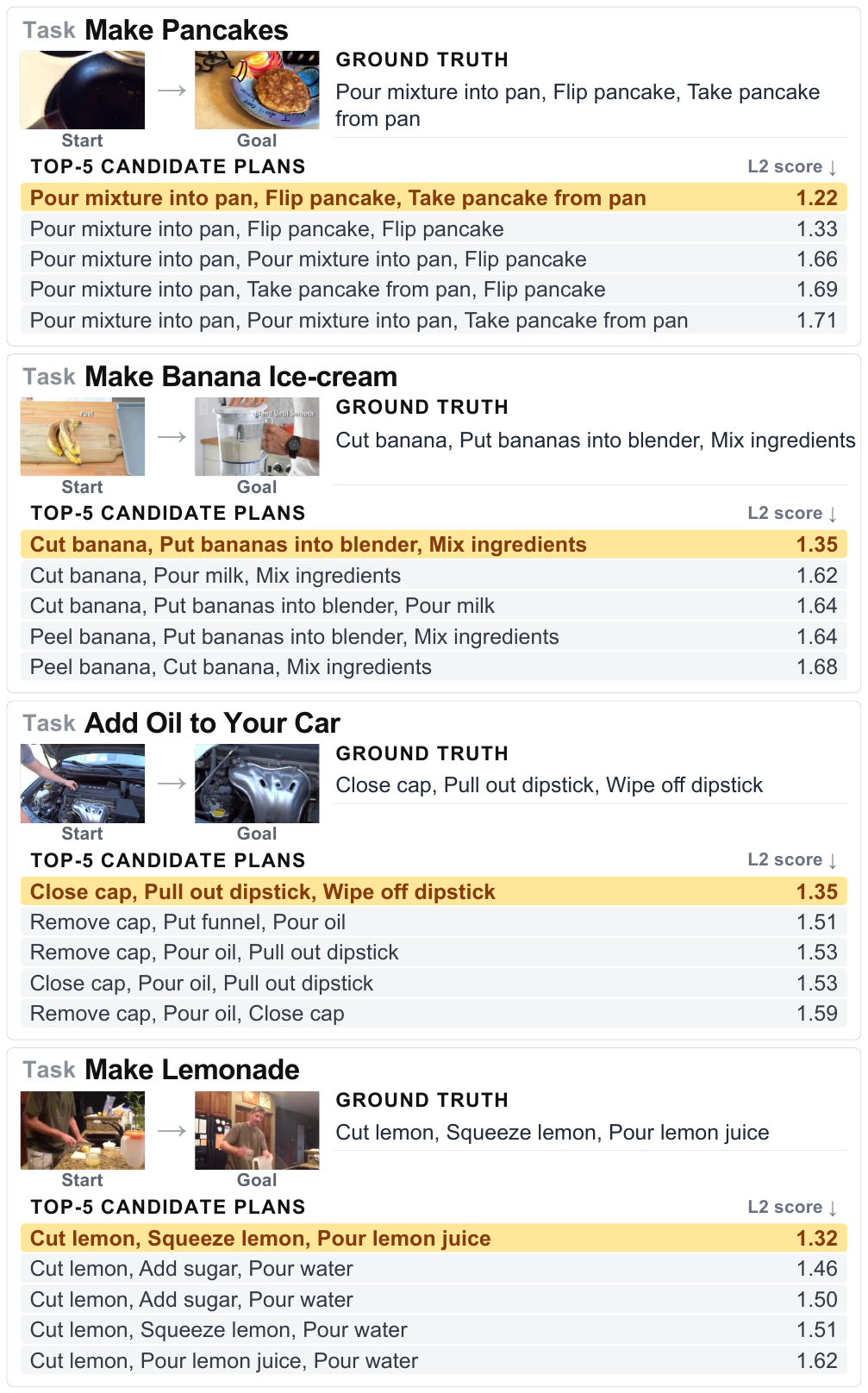}
    \caption{\textbf{Qualitative examples of \methodname{} on CrossTask.} Extension of \cref{fig:qualitative_success-1}.}   
    \label{fig:qualitative_success-2}
\end{figure}

In Figs.\ \ref{fig:qualitative_success-1} \& \ref{fig:qualitative_success-2}, we present qualitative examples of \methodname{} for procedure planning on \num{8} different tasks of the CrossTask dataset. In each example, the top-ranked candidate plan (\emph{i.e.}, the one with the lowest L2 distance to $x_{g}$) matches the ground truth action sequence. Still, all top-\num{5} candidates are plausible plans, but incorrect alternatives which involve step reordering (\emph{e.g.}, ``Flip pancake'' \emph{vs.} ``Take pancake from pan'' in ``Make pancakes''), action repetition (\emph{e.g.}, ``Flip steak'' in ``Grill steak''), and substitution with semantically related actions (\emph{e.g.}, ``Pour water'' \emph{vs.} ``Pour lemon juice'' in ``Make lemonade''). The modest margin between the correctly predicted plan and the alternatives suggests that the model can distinguish fine-grained semantics and temporal orderings. We emphasize that these examples are illustrative of correctly predicted plans selected for qualitative analysis.

\subsubsection{Failure cases.} 
\begin{figure}[t]
    \centering
    \includegraphics[width=0.875\linewidth]{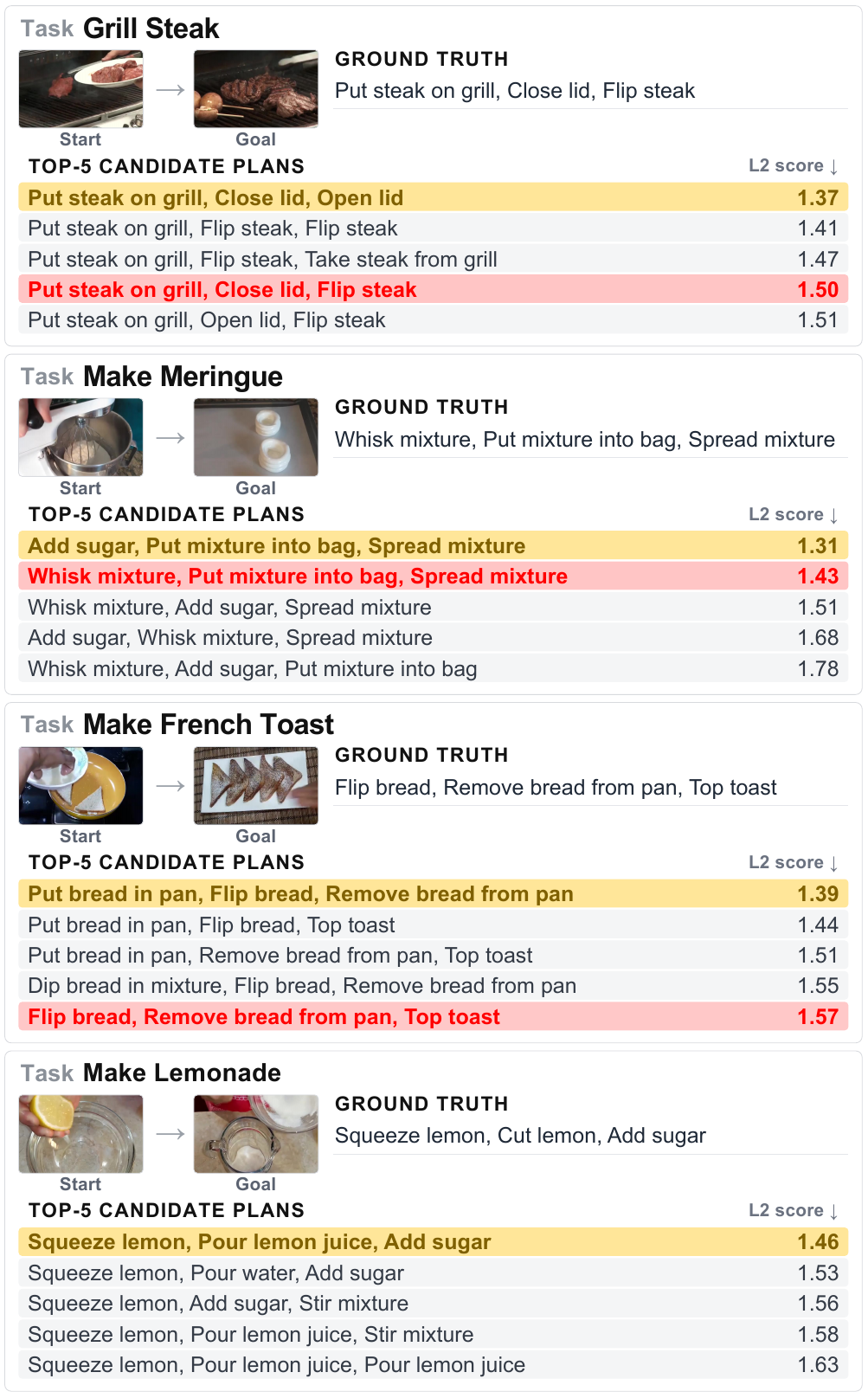}
    \caption{\textbf{Failure cases  of \methodname{} on CrossTask.} Procedural video results where the top-ranked candidate plan (highlighted in \colorindicator{myyellow}) fails to match the ground-truth action sequence (highlighted in \colorindicator{myred}). For some examples, the ground truth falls outside of the top-\num{5} candidates.}   
    \label{fig:qualitative_failure-1}
\end{figure}

\begin{figure}[t]
    \centering
    \includegraphics[width=0.875\linewidth]{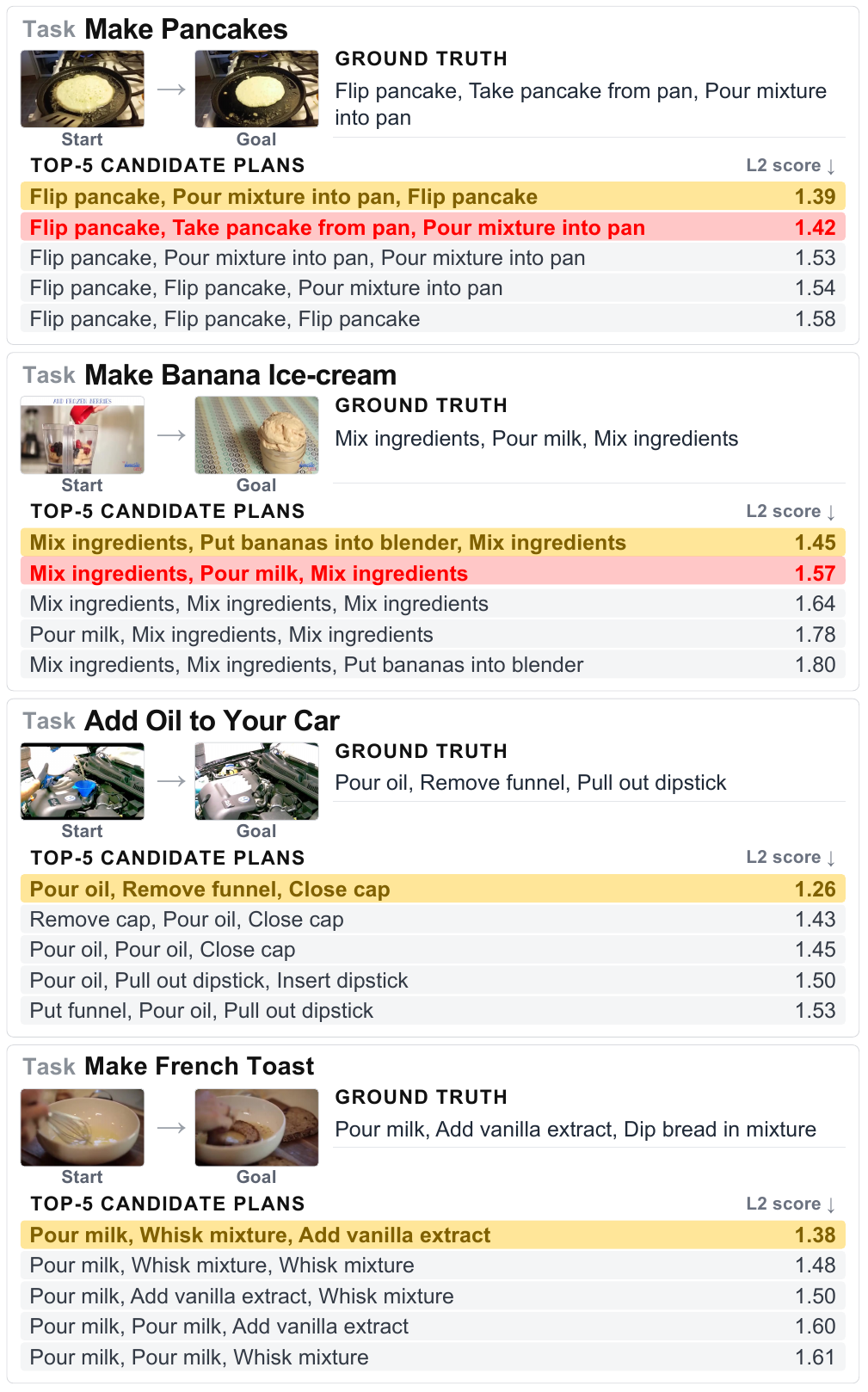}
    \caption{\textbf{Failure cases of \methodname{}  on CrossTask.} Extension of \cref{fig:qualitative_failure-1}.}   
    \label{fig:qualitative_failure-2}
\end{figure}

We show the failure examples of \methodname{} on the CrossTask dataset in Figs.\ \ref{fig:qualitative_failure-1} \& \ref{fig:qualitative_failure-2}. In some examples, the correct action sequence is not top-ranked but is still within the top-\num{5} action sequence candidates (\emph{e.g.}, ``Grill steak'', ``Make pancakes'', ``Make meringue''), while in others the models fail to infer the correct action sequence within the top-\num{5} sequence candidates. We identify several failure modes. In many wrong predictions, the model correctly identifies the first and the last action steps and fails to infer the intermediate action steps (\emph{e.g.}, ``Make lemonade'', ``Make pancakes'', ``Make banana ice-cream''). We attribute this to the fact that intermediate visual observation is not available, and the model has to rely solely on start and goal observations. Another observation is that the model wrongly predicts the first action step but successfully infers the subsequent steps (\emph{e.g.}, ``Make meringue'').

\end{document}